\documentclass[runningheads]{llncs}

\usepackage[eccv,year=2026,ID=12426]{eccv}

\usepackage{eccvabbrv}
\usepackage{graphicx}
\usepackage{booktabs}
\usepackage{cite}
\usepackage{amsmath}
\usepackage{bbm} 
\usepackage{amssymb}
\usepackage{multirow}
\usepackage{enumitem}
\usepackage{algorithm}
\usepackage{algpseudocode}
\usepackage{pifont}
\usepackage{caption}
\usepackage{colortbl}
\usepackage[table]{xcolor}
\usepackage[accsupp]{axessibility}
\definecolor{hdrgray}{RGB}{242,242,242}      % very light gray
\definecolor{baselineY}{RGB}{255,250,230}    % light warm yellow
\definecolor{tfY}{RGB}{255,242,204}          % slightly stronger yellow
\definecolor{oursMint}{RGB}{232,252,233}     % very light mint-green (best to highlight ours)

\usepackage{hyperref}
\usepackage{orcidlink}

\begin{document}

\title{Excite, Attend and Segment (EASe): Domain-Agnostic Fine-Grained Mask Discovery with Feature Calibration and Self-Supervised Upsampling}

\titlerunning{EASe: Domain-Agnostic Fine-Grained Mask Discovery}

\author{Deepank Singh\thanks{Corresponding author: \email{dksingh@uh.edu}} \and Anurag Nihal \and Vedhus Hoskere}
\authorrunning{D. Singh et al.}
\institute{University of Houston, Houston, TX, USA}

\maketitle

% ---------------------------------------------------------------
\begin{abstract}
\sloppy
Unsupervised segmentation approaches have increasingly leveraged foundation models (FM) to improve salient object discovery. However, these methods often falter in scenes with complex, multi-component morphologies, where fine-grained structural detail is indispensable. Many state-of-the-art unsupervised segmentation pipelines rely on mask discovery approaches that utilize coarse, patch-level representations. These coarse representations inherently suppress the fine-grained detail required to resolve such complex morphologies. To overcome this limitation, we propose \textbf{E}xcite, \textbf{A}ttend and \textbf{Se}gment (EASe), an unsupervised domain-agnostic semantic segmentation framework for \textit{easy }fine-grained mask discovery across challenging real-world scenes. EASe utilizes novel \textbf{S}emantic-\textbf{A}ware \textbf{U}psampling with \textbf{C}hannel \textbf{E}xcitation (SAUCE) to \textit{excite} low-resolution FM feature channels for selective calibration and \textit{attends} across spatially-encoded image and FM features to recover full-resolution semantic  representations. Finally, EASe \textit{segments} the aggregated features into multi-granularity masks using a novel training-free \textbf{C}ue-\textbf{A}ttentive \textbf{F}eature Aggr\textbf{e}gator (CAFE) which leverages SAUCE attention scores as a semantic grouping signal. EASe, together with SAUCE and CAFE, operate directly at pixel-level feature representations to enable accurate fine-grained dense semantic mask discovery. Our evaluation demonstrates superior performance of EASe over previous state-of-the-arts (SOTAs)  across major standard benchmarks and diverse datasets with complex morphologies. Code is available at \url{https://ease-project.github.io/}.
\end{abstract}

% ===============================================================
\section{Introduction}
\label{sec:intro}

% Accurate 
% Annotation 
% Most
% Most FM-based 
 % \textbf{E}xcite, \textbf{A}ttend and \textbf{Se}gment (EASe),nsemantic  bothat pixel-level 

Annotation costs escalate rapidly with increased fidelity of supervised semantic segmentation, especially when labels demand scarce domain expertise or pixel-accurate delineation of fine-grained structures \cite{shen2023survey}. This has motivated unsupervised semantic segmentation methods that reduce reliance on dense annotations \cite{unsamv1,yu2025unsamv2}, with foundation models (FM) further enabling the transfer of semantic features from large-scale pre-training \cite{sam_2023, sam_limit}. However, many FM-based unsupervised approaches still operate at coarse patch-token resolution \cite{cutler_maskcut,tokencut,dwc}, which can miss fine detail common in specialized domains (e.g., cracks and surface defects in infrastructure inspections \cite{hoskere2025unified, ccelik2025pixels,MALEPATI2025106469,hoskere2021physics}; cluttered indoor scenes \cite{opvocseg2023,caesar2018cocostuff} and overhead remote-sensing imagery for mapping and damage assessment \cite{varghese2025viewdelta, singh2023post, singh2025multiclass, singh2023climate}).

% Annotation costs escalate rapidly with increased fidelity of supervised semantic segmentation, especially when labels demand scarce domain expertise or pixel-accurate delineation of fine-grained structures \cite{shen2023survey}. These bottlenecks have motivated unsupervised semantic segmentation methods that reduce reliance on dense pixel-level annotations \cite{unsamv1,yu2025unsamv2}. More recently, foundation models (FM) have made this direction more feasible by providing transferable semantic features from large-scale pre-training \cite{sam_2023, sam_limit}. However, many unsupervised approaches use FM features only at coarse patch-token resolution to discover semantic masks \cite{cutler_maskcut,tokencut,dwc}, which can miss fine detail and struggle with complex morphologies common in a wide variety of specialized domains ranging from construction and civil infrastructure inspections (e.g., cracks and surface defects), to cluttered indoor spaces (e.g., kitchens and offices), and overhead remote-sensing imagery (e.g., satellite/aerial views for mapping and post-disaster damage assessment)

FM-based unsupervised methods typically follow an extract–cluster–refine recipe for salient object discovery, often losing fine semantic structure. These approaches \cite{cutler_maskcut,diffseg,couairon2024diffcut} extract low-resolution features from backbones such as DINO \cite{dino}, CLIP \cite{clip}, and Stable Diffusion \cite{stablediffusion}, then apply clustering or graph partitioning to produce coarse segments. At patch-level resolution, sub-token semantic cues are aggregated within a single token receptive field and often vanish in subsequent representations. Consequently, no clustering, graph cut, or partitioning step can recover cues that were never encoded. To sharpen boundaries in coarse masks, many approaches, including \cite{cutler_maskcut,diffseg,couairon2024diffcut}, add image-space refinement modules such as DenseCRF \cite{denscrf} and PAMR \cite{pamr}. These refinements promote spatial coherence by encouraging nearby pixels with similar texture to share labels. This texture-consistency assumption works for compact objects but can merge disconnected thin structures and fill gaps into surrounding regions, yielding inaccurate segmentation for fragmented or fine-grained morphological targets.

% FM-based unsupervised methods typically follow an extract, cluster, and refine recipe for salient object discovery, which often leads to a loss of fine semantic structure. These approaches \cite{cutler_maskcut,diffseg,couairon2024diffcut} extract low-resolution features from backbones such as DINO \cite{dino}, CLIP \cite{clip}, and Stable Diffusion \cite{stablediffusion}, then apply clustering or graph partitioning to produce coarse segments. At coarse patch-level resolution, sub-token semantic cues are aggregated within a single token receptive field and often disappear in the subsequent patch-level representations. Consequently, no subsequent clustering, graph cut, or partitioning step can recover the semantic cues that were never encoded. To sharpen boundaries in coarse masks, many approaches, including \cite{cutler_maskcut,diffseg,couairon2024diffcut}, incorporate image-space refinement modules such as DenseCRF \cite{denscrf} and PAMR \cite{pamr}. These refinements promote spatial coherence by encouraging nearby pixels with similar texture to share labels. Such texture-consistency assumption is effective for compact objects but often merge disconnected thin structures and fill gaps into surrounding regions, leading to inaccurate segmentation for fragmented or fine-grained morphological targets.

In this study, we present \textbf{E}xcite, \textbf{A}ttend and \textbf{Se}gment (EASe), an unsupervised semantic segmentation framework for \textit{easy} fine-grained mask discovery at high-resolution. Our approach builds on insights from recent SOTA methods to leverage self-supervised FM features for rich transferable semantics but overcomes the coarse-resolution feature bottleneck by attention-guided upsampling of FM features into dense pixel-level semantic representations. A training-free cue-guided aggregator uses the attention scores, learned during self-supervised upsampling as grouping signal to produce fine-grained semantic masks. Consequently, EASe preserves fine semantic structure and supports complete semantic understanding even in domains with complex and fragmented morphologies, as demonstrated in our results. 
The key contributions of our study are summarized as follows:
\begin{itemize}
    \item We introduce \textbf{Excite, Attend and Segment (EASe)}, a \textbf{unified framework} for unsupervised, domain-agnostic semantic segmentation that replaces coarse feature extraction, expensive graph cuts, and image-based refinement with a single scalable architecture operating at pixel resolution.
    \item We propose \textbf{SAUCE}, a \textbf{self-supervised upsampler} that augments cross-attention with channel calibration to amplify semantically discriminative dimensions when lifting patch tokens to pixel-level features.
    \item We propose \textbf{CAFE}, a \textbf{training-free aggregator} that repurposes learned attention of SAUCE as a grouping signal to produce coherent segmentations without additional training.
   \item We demonstrate through extensive experiments and ablation studies across 9 different standard and domain-specific benchmarks that \textbf{EASe consistently outperforms prior SOTA unsupervised segmentation} approaches with significant performance gains on fine-grained and morphologically complex targets.
\end{itemize}

% ===============================================================
\section{Related Works}
\label{sec:related}
\subsection{Unsupervised Semantic Segmentation}
\label{sec:rel_trainfree}
Recent unsupervised segmentation approaches \cite{lost,cutler_maskcut} leverage coarse-resolution FM features to broadly focus on salient object discovery, but struggle to preserve fine-grained semantics. Token leverages Normalized Cut ($N$-Cut) \cite{ncut} for single salient object localization while MaskCut \cite{cutler_maskcut} leverages iterative $N$-Cut to discover multiple object masks per image. CRF-based refinement remains common in \cite{tokencut,cutler_maskcut,lost} that assumes nearby pixels with similar texture should share labels, which often over-smooths thin structures and merges fragmented components. To exploit richer semantic priors, later methods such as \cite{dwc,diffseg,couairon2024diffcut} shifted to diffusion model features. Diffuse, Attend and Segment \cite{diffseg} utilizes self-attention layers of Stable Diffusion in a training-free iterative merging process based on KL divergence to produce segmentation masks. DiffCut \cite{couairon2024diffcut} further improves semantic granularity by applying recursive $N$-Cut on diffusion UNet encoder features to adaptively regulate segment count without a pre-determined object number but still relies on PAMR-based refinement \cite{pamr} to produce sharp masks. Current methods such as \cite{cutler_maskcut,diffseg,couairon2024diffcut} still operate on latent token representations where sub-token level structural semantics are often lost before downstream partitioning or refinement. 
% In our study, EASe overcomes coarse patch-level semantics by using SAUCE for high-resolution upsampled features and CAFE for training-free multi-granularity feature aggregation, enabling semantic grouping and mask discovery at pixel-level.
\subsection{Feature Upsampling for Foundation Models}
\label{sec:rel_upsample}
\sloppy Recent feature upsampling approaches for dense segmentation tasks covers learned content-aware modules and task-agnostic self-supervised methods, with distinct efficiency and generalization tradeoffs. Content-aware upsamplers such as CARAFE~\cite{wang2019carafe} and SAPA~\cite{lu2022sapa} predict adaptive kernels conditioned on local features, but rely on task-specific supervision, which limits transfer across downstream applications. Task-agnostic approaches such as LiFT~\cite{suri2024lift} and FeatUp~\cite{fu2024featup} remove task labels by using self-supervised proxy objectives. LiFT is limited to fixed-scale upsampling, while FeatUp can over-smooth features or require per-image inference-time optimization. JAFAR \cite{jafar} advances task-agnostic upsampling through lightweight cross-attention that promotes semantic alignment between high-resolution image-derived queries and low-resolution keys via a spatial feature transform module. JAFAR, however, treats all low-resolution feature channels as equally informative during upsampling. In domain transfer scenarios, channel importance shifts relative to the target domain and misalignment propagates through upsampling without explicit calibration. 
% We address this with SAUCE in EASe, which augments JAFAR with squeeze-and-excitation \cite{hu2018squeeze} to calibrate feature channels and consistently improves representations across backbones and domains.
% ===============================================================
\section{Excite, Attend and Segment (EASe) Methodology}
\label{sec:method}
% \vspace{-0.35in}
\begin{figure*}[h!]
  \centering
  \includegraphics[width=\linewidth]{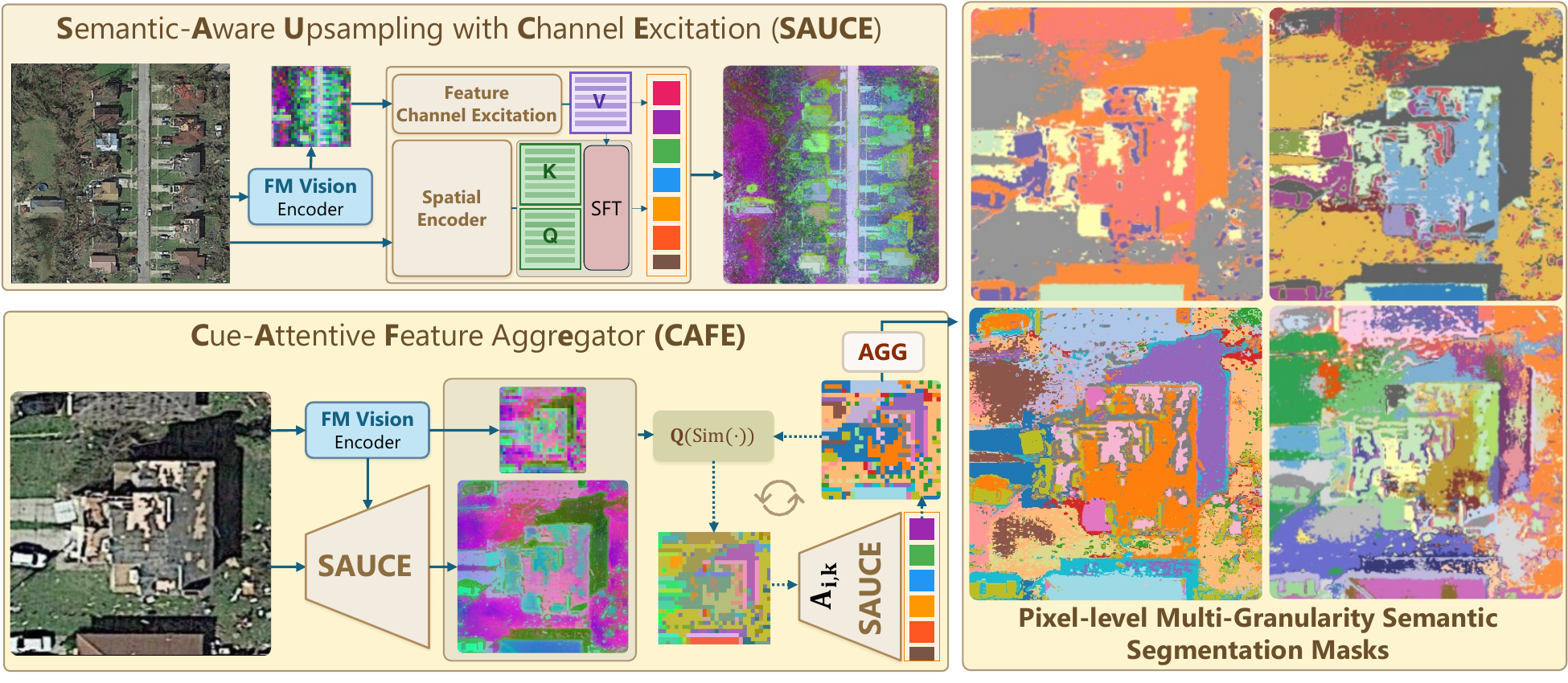}
  \caption{Overview of Excite, Attend and Segment (EASe). (Top) SAUCE (Semantic-Aware Upsampling with Channel Excitation) lifts coarse foundation model tokens to pixel-level semantic features via SE-calibrated cross-attention, where channel-excited features serve as both values and key modulators through SFT conditioning. (Bottom) EASe then utilises CAFE (Cue-Attentive Feature Aggregator), which first quantizes (Q) SAUCE's upsampled features into prototypes, then leverages SAUCE's to reduce and merge prototypes into coherent segments using Attention-Guided Grouping (AGG), followed by hierarchical agglomerative merging to produce pixel-level multi-granularity semantic segmentation masks (Right).}
  \label{fig:ease}
\end{figure*}
% ───────────────────────────────────────────────────────
% We present EASe: Excite, Attend and SEgment, an unsupervised semantic segmentation framework for fine-grained mask discovery in scenes with complex, multi-component morphology. Our method builds upon the insight that recent state-of-the-art methods utilise self-supervised FM features for mask discovery. However, these methods operate on coarse patch-level representations that average away fine structural detail, rely on pairwise affinity matrix  $\mathcal{O}(N^2)$ for spectral grouping that is already computationally expensive on coarse backbone features, and apply image-based spatial-coherence refinements that actively suppress the disconnected, thin, and heterogeneous structures that complex morphological scenes demand.
% Our propsed EASe addresses both limitations by operating directly at pixel-level feature representations to enable accurate fine-grained dense semantic mask discovery. The overall pipeline is illustrated in \cref{fig:ease-pipeline}. First SAUCE, a self-supervised cross-attention upsampler that lifts coarse patch-level backbone tokens to pixel-level semantic features via image-guided attention (\cref{sec:sauce}). Second, we introduce CAFE, a training-free cue-guided aggregator that repurposes SAUCE's learned cross-attention scores as a semantic grouping and filtering signal for fine-grained semantic mask discovery, eliminating the need for computationally expensive spectral clustering and image-based refinement. (\cref{sec:cafe}).
\noindent Excite, Attend and Segment (EASe) is an unsupervised semantic segmentation framework that operates at pixel-level resolution to perform fine-grained mask discovery in scenes with complex and multi-component morphology. Existing methods exploit self-supervised FM features for mask discovery yet lose fine structural detail at coarse patch-level representations, incur $\mathcal{O}(N^2)$ pairwise affinity matrices computation costs that are computationally prohibitive even at backbone resolution, and apply spatial coherence refinements that ignores fine-grained morphologies. EASe addresses all three limitations through two synergistic components, illustrated in Fig.~\ref{fig:ease}. SAUCE upsamples coarse patch-level backbone tokens to pixel-level semantic features through self-supervised cross-attention upsampling with channel excitation (\cref{sec:sauce}). CAFE repurposes these learned attention scores as a training-free semantic grouping signal for fine-grained mask discovery at multiple granularities without spectral clustering or image-based refinement (\cref{sec:cafe}). We describe the methodology details of each component in the following subsections.

\subsection{Semantic-Aware Upsampling with Channel Excitation (SAUCE)}
\label{sec:sauce}
\begin{figure}[htbp]
  \centering
  \includegraphics[width=\linewidth]{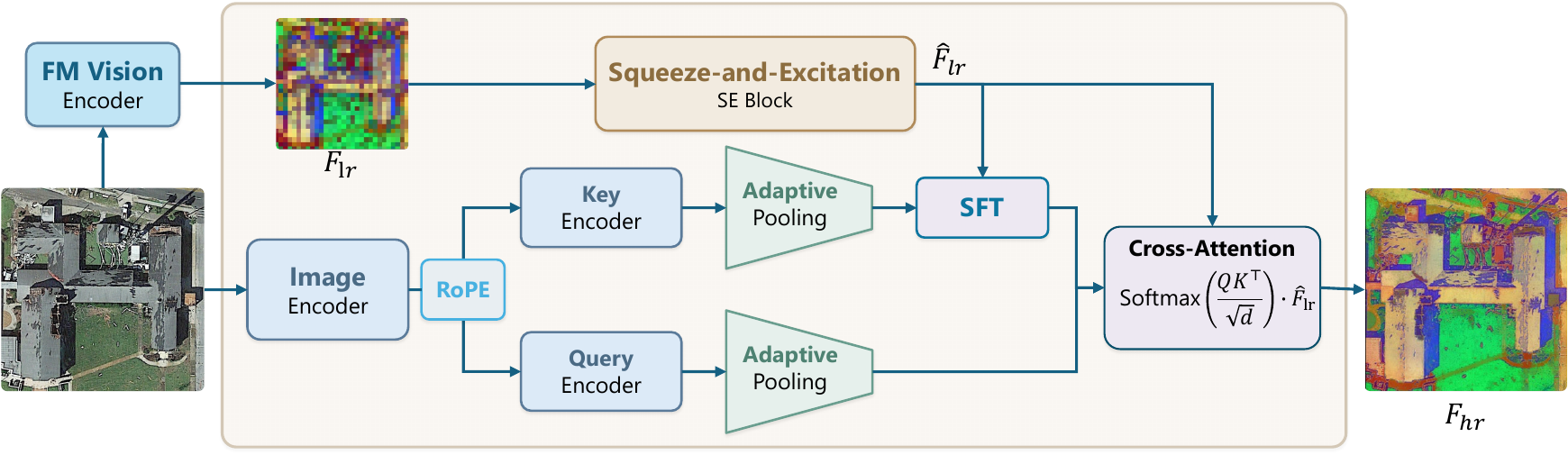}
    \caption{\textbf{SAUCE Architecture.} A frozen vision encoder extracts low-resolution tokens ($F_\text{lr}$). The \textbf{Channel Excitation} (Squeeze-and-Excitation (SE)) block recalibrates these tokens, which condition keys via Spatial Feature Transform (SFT) and serve directly as values in cross-attention. An image encoder with Rotary Position Embeddings (RoPE) produces queries at full resolution and keys at token resolution. The resulting attention map transfers semantic content from the recalibrated tokens to every pixel, yielding the upsampled feature map ($F_\text{hr}$).}
  \label{fig:sauce}
\end{figure}
\noindent SAUCE is a feature upsampler that uses the input image as high-resolution guidance to construct dense feature representations from coarse FM vision encoder features. Inspired by JAFAR~\cite{jafar}, SAUCE encodes the image through a lightweight encoder and applies RoPE positional embeddings to decouple the learned attention pattern from any fixed spatial grid. However, high-dimensional feature spaces contain many redundant channels that dilute the attention signal and weaken the semantics of upsampled features. SAUCE addresses this with a Squeeze-and-Excitation (SE)~\cite{hu2018squeeze} block to calibrate $\mathbf{F}_{\text{lr}}$ channels before upsampling, selectively amplifying salient dimensions through Eq. \ref{eq:se}.
\begin{equation}
\hat{\mathbf{F}}_{\text{lr}} = \mathbf{F}_{\text{lr}} \odot \sigma\Big(\mathbf{W}_2 , \operatorname{SiLU}\big(\mathbf{W}_1 , \operatorname{GAP}(\mathbf{F}_{\text{lr}})\big)\Big),
\label{eq:se}
\end{equation}
where $\operatorname{GAP}$ is global average pooling, $\mathbf{W}_1{\in}\mathbb{R}^{\frac{C}{2} \times C}$, $\mathbf{W}_2{\in}\mathbb{R}^{C \times \frac{C}{2}}$ and $\sigma$ is sigmoid. From the shared image representation, queries are pooled to the target output resolution while keys are downsampled to the encoder patch grid and modulated via Spatial Feature Transform (SFT) conditioned on $\hat{\mathbf{F}}_{\text{lr}}$. Cross-attention is computed with $\hat{\mathbf{F}}_{\text{lr}}$ as values, so the attention map directly interpolates recalibrated backbone features to every pixel, producing the full-resolution feature map $\mathbf{F}_{\text{hr}}$.

\smallskip\noindent\textbf{Training.}
SAUCE is trained in a self-supervised manner solely on multi-resolution image pairs using a combined cosine and MSE loss (Eq. \ref{eq:sauce_loss}). Given an image $\mathbf{I}$ and a downsampled view $\mathbf{I}_\text{lr}$, the model produces $\hat{\mathbf{F}}_\text{hr} = \textsc{sauce}(\mathbf{I},\, \mathbf{F}_\text{lr})$ to reconstruct the target $\mathbf{F}_\text{hr}{=}f(\mathbf{I})$ from $\mathbf{F}_\text{lr}{=}f(\mathbf{I}_\text{lr})$, where $f$ is a frozen backbone.
\begin{equation}
  \mathcal{L} = 1 - \cos(\hat{\mathbf{F}}_\text{hr},\, \mathbf{F}_\text{hr})
              + \|\hat{\mathbf{F}}_\text{hr} - \mathbf{F}_\text{hr}\|_2^2.
  \label{eq:sauce_loss}
\end{equation}

% ───────────────────────────────────────────────────────
\subsection{Cue-Attentive Feature Aggregator (CAFE)}
\label{sec:cafe}

\begin{figure}[t]
  \centering
  \includegraphics[width=\linewidth]{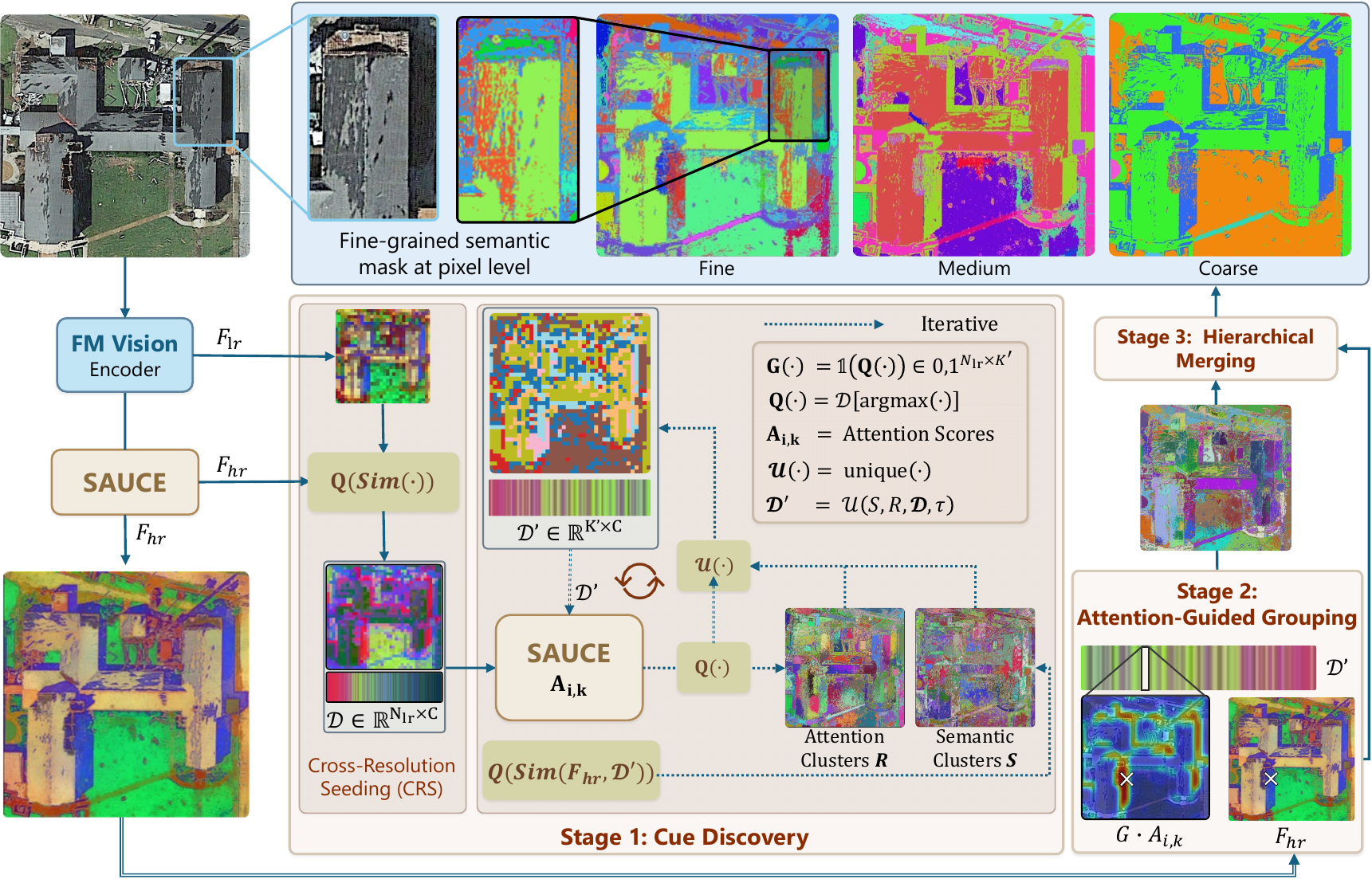}
  \caption{\textbf{Overview of the CAFE architecture.} In CAFE, low-resolution features $F_\text{lr}$ from the FM vision encoder are passed through SAUCE, which outputs attention scores $\mathbf{A}_{i,k}$ alongside $F_{hr}$. Stage 1 (Cue Discovery via Cross-Resolution Seeding, CRS) initializes a prototype dictionary $\mathcal{D} \in \mathbb{R}^{N_{lr} \times C}$ from quantized $F_\text{lr}$ and $F_\text{hr}$. $\mathcal{U}\left(\cdot\right)$ operator merges quantized attention clusters $\mathbf{R}$ and semantic clusters $\mathbf{S}$ to obtain refined prototypes $\mathcal{D}' \in \mathbb{R}^{K' \times C}$. $\mathcal{D}'$ is iteratively refined in Stage 1 until convergence. Stage 2 (Attention-Guided Grouping, AGG) assigns every pixel a label by combining the grouped affinity map $G \cdot A_{i,k}$ with $F_\text{hr}$ under the final $\mathcal{D}'$. Stage 3 (Hierarchical Merging, HM) progressively merges dense pixel-level assignments into coarser partitions. The resulting fine-to-coarse segmentation hierarchy spanning various fine-to-coarse levels is shown in the top panel.}
  \label{fig:cafe}
\end{figure}
CAFE is a training-free semantic grouping module that operates in pixel-level features ${F}_{\text{hr}}$ by reusing SAUCE cross-attention $\mathbf{A}_{i,k}$ between high-resolution queries and low-resolution keys ${F}{\text{lr}}$ as summarized in Fig.~\ref{fig:cafe}. Unlike object-focused mask discovery, it emphasizes semantic detail by converting the pre-trained, continuous attention grouping into discrete segment assignments. In Stage 1, Cue Discovery via CRS constructs a compact prototype dictionary $\mathcal{D}$ with $K$ cues and prunes redundancies to obtain $\mathcal{D}'$. In Stage 2, Attention-Guided Grouping uses $\mathbf{A}_{i,k}$ to project $\mathcal{D}'$ onto ${F}_{\text{hr}}$ and produce $K$ dense affinity maps for semantic clustering. This design removes the pixel-level $\mathcal{O}(N^2)$ affinity bottleneck and avoids recursive partitioning or image-space refinement that can erase thin or fragmented semantics. Stage 3 Hierarchical Merging consolidates the discovered semantic clusters and progressively organizes them into fine, medium, and coarse masks. The next subsection details Stage 1 Cue Discovery via CRS, Stage 2 Attention-Guided Grouping, and Stage 3 Hierarchical Merging.

\paragraph{\textbf{Stage 1: Cue Discovery via Cross-Resolution Seeding (CRS).}}
The initial prototype dictionary $\mathcal{D} \in \mathbb{R}^{N_{\text{lr}} \times C}$ is obtained via cross-resolution seeding, mapping each low-resolution spatial token to its nearest high-resolution feature under cosine similarity. We define the quantizer $\mathbf{Q}(\cdot) = \mathcal{D}[\arg\max(\cdot)]$, which assigns each pixel to its nearest prototype. CAFE iteratively refines $\mathcal{D}$ through two complementary quantized signals, \textit{(i) attention clusters} $R = \mathbf{Q}(\mathbf{A}_{i,k})$, where SAUCE attention scores, $\mathbf{A}_{i,k}$  are quantized to assign each pixel to its dominant prototype, proposing coarse spatial groups and \textit{(ii) semantic clusters} $S = \mathbf{Q}(\mathrm{Sim}(\mathbf{F}_{\text{hr}}, \mathcal{D}))$, where cosine similarity over high-resolution features assigns each pixel to its nearest prototype, acting as a fine-grained semantic gate. The merge operator $\mathcal{U}(S, R, \mathcal{D}, \tau)$ cross-tabulates $S$ and $R$ to map each semantic cluster to its dominant attention group, then within each group merges prototypes whose pairwise cosine similarity exceeds $\tau$ by averaging, while discarding prototypes that receive no pixel assignment under $S$. This distills $\mathcal{D}$ to a compact $\mathcal{D}' \in \mathbb{R}^{K' \times C}$ with $K' \ll N_{\text{lr}}$. The process iterates, replacing $\mathcal{D} \leftarrow \mathcal{D}'$, until $|\mathcal{D}'| \geq |\mathcal{D}|$, retaining only prototypes that are spatially grounded by attention and semantically distinct within their group.
\paragraph{\textbf{Stage 2: Attention-Guided Grouping (AGG).}}
To obtain dense semantic clusters from the final prototype dictionary from Stage~1, $\mathcal{D}'$ with $K'$ prototypes, we assign every high-resolution pixel a label by minimising a blended cost by combining feature similarity with spatial attention affinity (Fig.~\ref{fig:cafe}, Stage~2). The one-hot grouping matrix $\mathbf{G}(\cdot) = \mathbbm{1}(\mathbf{Q}(\cdot)) \in \{0,1\}^{N_{\text{lr}} \times K'}$ assigns each LR token to its nearest prototype in $\mathcal{D}'$. The attention scores $\mathbf{A}_{i,k}$ are then pooled per prototype as $\mathbf{A}^m = \mathbf{A}_{i,k} \cdot \mathbf{G} \in \mathbb{R}^{N_{\text{hr}} \times K'}$, yielding $K'$ per-pixel affinity maps. For $\ell_2$-normalised HR feature $\mathbf{f}_i \in \mathbb{R}^{C}$ at pixel $i \in \{1, \dots, N_{\text{hr}}\}$ and $\ell_2$-normalised prototype $\mathbf{d}_k \in \mathbb{R}^{C}$, $k \in \{1, \dots, K'\}$, the feature and affinity costs are defined respectively as
\begin{equation}
  \mathcal{C}^{\text{feat}}_{i,k} = 1 - \mathbf{f}_i^\top \mathbf{d}_k, \qquad
  \mathcal{C}^{\text{aff}}_{i,k}  = 1 - (A^m)_{i,k},
\end{equation}
where $(A^m)_{i,k} \in [0,1]$ is the total attention affinity of HR pixel $i$ to prototype $k$. The final per-pixel assignment is
\begin{equation}
  y_i = \arg\min_k \; \alpha\,\mathcal{C}^{\text{feat}}_{i,k} + \beta\,\mathcal{C}^{\text{aff}}_{i,k}.
\end{equation}
The resulting label map is further refined through edge-aware boundary smoothing, where pixels surrounded by a majority of a neighbouring label are reassigned unless they lie on a strong feature gradient.

\paragraph{\textbf{Stage 3: Hierarchical Merging (HM).}}
% The final label map from Stage 2 is highly fine-grained, containing up to $K'$ clusters. To recover semantically meaningful partitions without supervision, we perform agglomerative merging by sweeping a cosine similarity threshold from $\tau$ to 0.3 with a smaller step-size of 0.001. At each threshold, clusters with a mean pair-wise prototype cosine similarity exceeding the current threshold are merged by averaging, resulting in a sequence of progressively coarser label maps. First, all resulting levels are ranked by the Calinski–Harabasz (CH) index~\cite{Caliński01011974} computed as the ratio of between-cluster to within-cluster variance in L2-normalized high-resolution feature space, and the top-$N$ levels with the highest CH scores are retained, yielding a multi-granularity hierarchy $\{L_0, \dots, L_{N-1}\}$ ordered from fine to coarse. Then at each retained level, we compute two image-level granularity scores adapted from GraCo~\cite{zhao2024gracogranularitycontrollableinteractivesegmentation}, (i) \emph{Scale granularity} $G_{\text{scale}}$, defined as the mean ratio of cluster area to foreground area, quantifying how much of the image each cluster occupies, and (ii) \emph{Semantic granularity} $G_{\text{sem}}$, defined as the mean ratio of within-cluster to global feature spread, quantifying the semantic discriminability of each cluster relative to the full scene. These metrics enable automatic level selection based on average target annotation granularity for evaluation.

Fine label map, obtained from Stage~2, contains upto $K'$ clusters (Fig.~\ref{fig:cafe}). Stage~3 applies agglomerative merging by sweeping a cosine-similarity threshold from $\tau$ down to $0.3$ in steps of $0.001$, and merges clusters whose mean prototype similarity exceeds the current threshold. This generates a sequence of label maps in fine to coarse order. We rank all candidate levels using the Calinski–Harabasz (CH) index~\cite{Caliński01011974} computed in L2-normalized high-resolution feature space, and keep the top-$N$ levels as a hierarchy ${L_0,\dots,L_{N-1}}$. We compute two image-level granularity scores for each retained level and are adapted from GraCo~\cite{zhao2024gracogranularitycontrollableinteractivesegmentation} as $G_{\text{scale}}$ (mean cluster-area over foreground-area) and $G_{\text{sem}}$ (mean within-cluster spread over global spread). These scores enable automatic level selection by matching the target annotation granularity during evaluation, without supervision.

\paragraph{\textbf{Granularity Selection and Evaluation.}} Prior training-free methods cut subgraphs of an affinity matrix in a coarse-to-fine recursion depth~\cite{yu2025unsamv2,couairon2024diffcut}, which looses global context and requires dataset-specific recursion depth. CAFE produces fine-to-coarse multi-granularity masks in one forward pass, with no recomputation or parameter adjustment. For evaluation, the same pool of candidate levels serves every benchmark without re-extraction. For instance, COCO-Stuff-27 and COCO-Object-80 select different levels from the same multi-granularity candidates, where a small subset from the training set determines the target $(G_{\text{scale}}, G_{\text{sem}})$ and the level closest to this target is selected.

\section{Experimentation details}\label{sec:expdet}
\subsection{Datasets and Metrics}
Following the existing works \cite{cho2021picie,zhou2022extract,shin2022reco,li2023acseg,couairon2024diffcut}, we evaluate across different datasets to cover diverse object categories, image styles, resolutions, and viewpoints. Standard benchmarks including Pascal VOC \cite{everingham2015pascal} (20 foreground classes), COCO-Object \cite{cocolin2014microsoft} (80 foreground classes), COCO-Stuff \cite{caesar2018cocostuff} which merges 80 things and 91 stuff categories in COCO-Stuff into 27 mid-level classes, Cityscapes \cite{cordts2016cityscapes} (27 foreground classes), KITTI \cite{kitti} (27 foreground classes, same as Cityscapes), PartImageNet \cite{partimgnet} (40 classes including background) and ADE20K \cite{ade20k} (150 foreground classes) are used for evaluation in our study. Pascal VOC and COCO-Object include an additional background class. Evaluation is performed on original validation splits for all datasets except COCO and KITTI. For COCO dataset, we adopt the validation split established by prior works \cite{cho2021picie,chopicie,ji2019invariant,ji51804invariant} and followed in \cite{couairon2024diffcut}. For KITTI dataset, we directly evaluate on the original training set due to unavailability of validation set and small size of training set. We additionally evaluate on OmniCrack30k \cite{benz2024omnicrack30k} (binary class) and Roof Subassembly Damage Detection \cite{roofdamage,roofdamagepaper} (8 classes including background) as specialized benchmarks with fine-grained and fragmented morphological structures to assess EASe as a universal unsupervised segmentation framework.

For all datasets, we report the mean intersection over union \cite{iou} (mIoU) which is the most popular evaluation metric for semantic segmentation. Following \cite{couairon2024diffcut}, predicted masks are assigned to ground truth via Hungarian matching \cite{hungarianmatching}, with many-to-one matching for the background class across Pascal VOC, COCO-Object, PartImageNet, OmniCrack30k, and Roof Subassembly Damage Detection datasets.

\subsection{Implementation Details}\label{sec:impl}
% \vspace{-0.5in}
EASe uses a DINOv3~\cite{dino} ViT-L/16 FM backbone. For all unsupervised semantic segmentation experiments, SAUCE is self-supervisedly trained on ImageNet without labels for 127,500 steps using the AdamW optimizer with a learning rate of $2\times10^{-4}$. All CAFE hyperparameters are fixed across datasets, images are resized to $640$ px on the longest side while preserving the aspect ratio, Stage-1 Cue Discovery $\tau{=}0.97$ is used. For stage-2, Attention-Guided Grouping, $\alpha{=}0.75$ and $\beta{=}0.25$ are set. Stage-3, Hierarchical Merging ($\theta \in [0.99, 0.30]$ with $\delta{=}1{\times}10^{-3}$). As annotation granularity varies across target datasets, we follow prior work~\cite{diffseg,couairon2024diffcut} to adapting to the inherent granular complexity of each domain. We calibrate the granularity scores, $G_{\text{scale}}$ and $G_{\text{sem}}$, on a small subset of annotated images, taking the smaller of 200 images or 10\% of the target training split. The calibrated values for each dataset are provided in the supplementary material. 

\section{Results and Discussion}
\begin{table}[ht]
\centering
\caption{\textbf{Comparisons for unsupervised semantic segmentations.} We compare EASe to SOTA methods in unsupervised segmentation tasks on Pascal VOC \cite{everingham2015pascal}, COCO-Object (COCO-O) \cite{cocolin2014microsoft}, COCO-Stuff (COCO-S) \cite{caesar2018cocostuff}, Cityscapes \cite{cordts2016cityscapes}, ADE20K \cite{ade20k}, PartImageNet (PIN) \cite{partimgnet}, KITTI \cite{kitti}, OmniCrack30K (OC30K) \cite{benz2024omnicrack30k}, and Roof Subassembly Damage Detection (RSDD) \cite{roofdamage} datasets. EASe achieves state-of-the-art results on 8 out of 9 benchmarks, outperforming competitive approaches by an average of +3.7 mIoU on seven standard segmentation benchmarks and +19.8 mIoU on datasets with complex morphology. Best results are in \textbf{bold}, second best results are \underline{underlined}.}
\label{tab:main}
\resizebox{\linewidth}{!}{%
\setlength{\tabcolsep}{4pt}
\begin{tabular}{lcccccccccc}
\toprule
\rowcolor{hdrgray}
\textit{Methods} & \textit{tf} & \textit{VOC} & \textit{COCO-O} & \textit{COCO-S} & \textit{Cityscapes} & \textit{ADE20K} & \textit{PIN} & \textit{KITTI} & \textit{OC30K} & \textit{RSDD} \\
\rowcolor{hdrgray}
% & & 20 & Obj-80 & Stuff-27 & 27 & 20K & ImgNet & &  &  \\
\midrule

% --- 2nd-5th rows (baselines): light yellow ---
\rowcolor{baselineY}
ReCO~\cite{shin2022reco}        & \ding{55} & 25.1 &  15.7 &  --  & 19.3 & 11.2 &  --  &  --  & -- & -- \\
\rowcolor{baselineY}
MaskCLIP~\cite{maskclip}        & \ding{55} & 38.8 &  20.6 &  --  & 10.0 &  9.8 &  --  &  --  & -- & -- \\
\rowcolor{baselineY}
STEGO~\cite{stego}              & \ding{55} &  --  &   --  & 28.2 & 21.0 &  --  &  --  &  --  & -- & -- \\
\rowcolor{baselineY}
ACSeg~\cite{li2023acseg}        & \ding{55} & 53.9 &   --  & 28.1 &  --  &  --  &  --  &  --  & -- & -- \\
\midrule

% --- 6th-8th rows (TF methods): stronger yellow ---
\rowcolor{tfY}
MaskCut ($k=5$)~\cite{cutler_maskcut} & \ding{51} & 53.8 & 30.1 & 41.7 & 18.7 & 35.7 &  38.4  &  11.8  & 9.4 & \underline{28.1} \\
\rowcolor{tfY}
DiffSeg~\cite{diffseg}                  & \ding{51} & 49.8 & 23.2 & 44.2 & 21.2 & 37.7 &  37.0  &  28.5  & \underline{45.0} & 21.7 \\
\rowcolor{tfY}
DiffCut~\cite{couairon2024diffcut}      & \ding{51} & \textbf{65.2} & \underline{34.1} & \underline{49.1} & \textbf{30.6} & \underline{44.3} & \underline{39.0} & \underline{34.0} & 35.4 & 21.2 \\
\midrule

% --- Ours: mint-green highlight (recommended) ---
\rowcolor{oursMint}
EASe (Ours) & \ding{51} & \underline{63.6} & \textbf{43.2} & \textbf{50.9} & \textbf{32.8} & \textbf{49.4} & \textbf{46.4} & \textbf{36.0} & \textbf{54.2} & \textbf{58.4} \\
\rowcolor{oursMint}
$\Delta$ \textit{prev.\ SOTA} & & \textcolor{black}{$-$1.6} & \textcolor{green!60!black}{$+$9.1} & \textcolor{green!60!black}{$+$1.8} & \textcolor{green!60!black}{$+$2.2} & \textcolor{green!60!black}{$+$5.1} & \textcolor{green!60!black}{$+$7.4} & \textcolor{green!60!black}{$+$2.0} & \textcolor{green!60!black}{$+$9.2} & \textcolor{green!60!black}{$+$30.3} \\
\bottomrule
\end{tabular}%

}% end resizebox
\end{table}
\label{sec:results}
\begin{figure}
    \centering
    \includegraphics[width=1\linewidth]{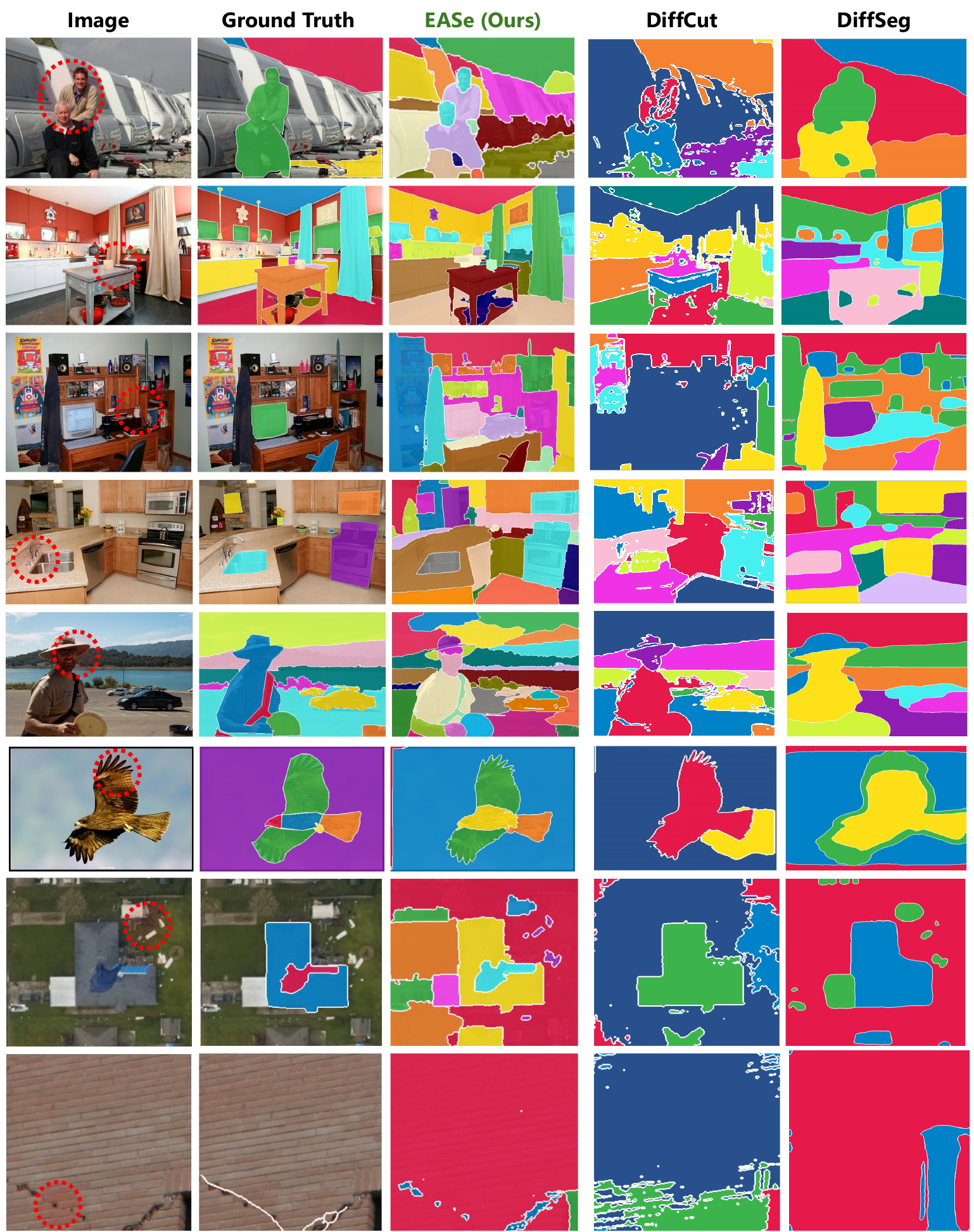}
    \caption{Qualitative comparison across datasets, with highlighted red-encircled region of interest. EASe consistently produces sharper, semantically correct masks, separating adjacent objects. Row 1 and Row 6 examples show global semantics preservation with heads, and feather correctly assigned to same respective semantic classes. In other cases, such as Rows 2-3, EASe distinguishes small items (e.g. boxes, bottles) on table. Examples in Rows 7-8 correctly capture damage pattern and fragmented object of debris precisely at pixel level, whereas DiffCut and DiffSeg either merge or miss them. 
}
    \label{fig:main_results}
\end{figure}
\noindent We evaluate EASe against training-free and trained unsupervised semantic segmentation methods across nine benchmarks spanning object-centric, scene-level, part-level, and domain-specific datasets. The results in Table~\ref{tab:main} demonstrate the three core advantages of EASe. First, operating at full pixel resolution with dense features enables consistent gains across increasingly complex label spaces, with margins reaching +9.1 mIoU on COCO-Object (81 classes) and +5.1 on ADE20K (150 classes), where previous methods struggle to distinguish densely overlapping categories. Second, preserving global semantic context in a fine-to-coarse hierarchy proves essential for part-level parsing, where EASe outperforms the strongest baseline by +7.4 on PartImageNet, capturing fine semantic variations within objects. Third, EASe surpasses all baselines by +9.2 on OmniCrack30K and +30.3 on RSDD with 7 damage categories, where fine-grained structures such as crack propagation patterns and roof subassembly damage demand sub-pixel semantic discrimination. Prior methods, built on object-level saliency priors, struggle most severely in these domains as their patch-resolution representations cannot resolve the complex morphology that EASe easily captures by operating in pixel-level feature space. The effectiveness of our framework is further illustrated in Fig.~\ref{fig:main_results}, where EASe produces semantically coherent masks with precise boundaries across all domains, while competing methods exhibit fragmented or over-merged regions. Additional qualitative examples are provided in the supplementary material.
\subsection{Ablation study}
\subsubsection{Upsampler Ablation}
\begin{table}[htbp]
\centering

\begin{minipage}[t]{0.49\textwidth}
\centering
\captionof{table}{\textbf{Linear probing (LP) evaluation of upsampler variants.}
We compare the current SOTA feature upsampler, JAFAR~\cite{jafar}, against our SAUCE upsampler under two configurations SE-Block frozen and SE-Block trainable during LP. Best results per backbone are \textbf{bold}, second best are \underline{underlined}.}
\label{tab:lp_ablation}
\resizebox{\linewidth}{!}{%
\setlength{\tabcolsep}{4pt}
\begin{tabular}{llcccc}
\toprule
\rowcolor{hdrgray}
\textit{Methods} & \textit{Encoder} & \textit{SE} & \textit{VOC} & \textit{COCO-Stuff} & \textit{ADE20K} \\
\midrule
\rowcolor{baselineY}
JAFAR~\cite{jafar}          & ViT-S  & --         & \underline{80.71} & \underline{59.61} & \underline{40.45} \\
\rowcolor{oursMint}
SAUCE (Ours)                & ViT-S  & \ding{55}  & 80.42 & 59.58 & \underline{40.45} \\
\rowcolor{oursMint}
SAUCE (Ours)                & ViT-S  & \ding{51}  & \textbf{81.40} & \textbf{61.19} & \textbf{40.74} \\
\rowcolor{oursMint}
$\Delta$ \textit{prev.\ SOTA} & & & \textcolor{green!60!black}{$+$0.69} & \textcolor{green!60!black}{$+$1.58} & \textcolor{green!60!black}{$+$0.29} \\
\midrule
\rowcolor{baselineY}
JAFAR~\cite{jafar}          & ViT-L & --         & 86.37 & 64.49 & 48.55 \\
\rowcolor{oursMint}
SAUCE (Ours)                & ViT-L & \ding{55}  & \underline{86.60} & \underline{64.62} & \underline{48.69} \\
\rowcolor{oursMint}
SAUCE (Ours)                & ViT-L & \ding{51}  & \textbf{87.60} & \textbf{65.65} & \textbf{49.09} \\
\rowcolor{oursMint}
$\Delta$ \textit{prev.\ SOTA} & & & \textcolor{green!60!black}{$+$1.23} & \textcolor{green!60!black}{$+$1.16} & \textcolor{green!60!black}{$+$0.54} \\
\bottomrule
\end{tabular}%
}
\end{minipage}
\hfill
\begin{minipage}[t]{0.49\textwidth}
\centering
\captionof{table}{\textbf{Upsampler reconstruction quality.}
We compare the feature reconstruction fidelity of JAFAR~\cite{jafar} and our SAUCE upsampler by measuring MSE ($\times 10^{3}$, $\downarrow$) and Cosine Similarity ($\uparrow$) between upsampled and ground-truth high-resolution backbone features on three datasets. Best results are \textbf{bold}.}
\label{tab:upsampler_quality}
\resizebox{\linewidth}{!}{%
\setlength{\tabcolsep}{4pt}
\begin{tabular}{llcccccc}
\toprule
\rowcolor{hdrgray}
& & \multicolumn{2}{c}{\textit{VOC}} & \multicolumn{2}{c}{\textit{COCO}} & \multicolumn{2}{c}{\textit{ADE20K}} \\
\cmidrule(lr){3-4} \cmidrule(lr){5-6} \cmidrule(lr){7-8}
\rowcolor{hdrgray}
\textit{Methods} & \textit{Encoder} & MSE $\downarrow$ & Sim $\uparrow$ & MSE $\downarrow$ & Sim $\uparrow$ & MSE $\downarrow$ & Sim $\uparrow$ \\
\midrule
\rowcolor{baselineY}
JAFAR~\cite{jafar}  & ViT-S  & 2.12 & 0.884 & 2.20 & 0.877 & 2.07 & 0.885 \\
\rowcolor{oursMint}
SAUCE (Ours)        & ViT-S  & \textbf{2.06} & \textbf{0.893} & \textbf{2.11} & \textbf{0.886} & \textbf{2.03} & \textbf{0.893} \\
\rowcolor{oursMint}
$\Delta$ \textit{prev.\ SOTA} & & \textcolor{green!60!black}{$-$2.8\%} & \textcolor{green!60!black}{$+$1.0\%} & \textcolor{green!60!black}{$-$4.1\%} & \textcolor{green!60!black}{$+$1.0\%} & \textcolor{green!60!black}{$-$1.9\%} & \textcolor{green!60!black}{$+$0.9\%} \\
\midrule
\rowcolor{baselineY}
JAFAR~\cite{jafar}  & ViT-L & 1.62 & 0.854 & 1.74 & 0.845 & 2.15 & 0.824 \\
\rowcolor{oursMint}
SAUCE (Ours)        & ViT-L & \textbf{1.56} & \textbf{0.867} & \textbf{1.66} & \textbf{0.859} & \textbf{2.03} & \textbf{0.841} \\
\rowcolor{oursMint}
$\Delta$ \textit{prev.\ SOTA} & & \textcolor{green!60!black}{$-$3.7\%} & \textcolor{green!60!black}{$+$1.5\%} & \textcolor{green!60!black}{$-$4.6\%} & \textcolor{green!60!black}{$+$1.7\%} & \textcolor{green!60!black}{$-$5.6\%} & \textcolor{green!60!black}{$+$2.1\%} \\
\bottomrule
\end{tabular}%
}
\end{minipage}

\end{table}
\sloppy
We follow the linear probing protocol of~\cite{jafar}, training a linear segmentation head on frozen upsampler weights pretrained self-supervised for 127.5K steps. To isolate the SE block's contribution, we evaluate two configurations: (i)~SE frozen, where the full upsampler is pretrained on the target dataset and SE gates remain fixed during probing; and (ii)~SE trainable, where the upsampler is pretrained on ImageNet and only the SE gates adapt during probing. In \cref{tab:lp_ablation}, for ViT-S (384-d), freezing SE yields performance on par with JAFAR, as the lower-dimensional feature space offers limited redundancy for channel recalibration to exploit. For ViT-L (1024-d), frozen SE already gains +0.23 mIoU on VOC and +0.14 on ADE20K, indicating that the higher-dimensional space enables the channel gates to learn meaningful importance priors during self-supervised pretraining. Unfreezing SE during probing amplifies gains across both backbones, reaching +1.58 mIoU on COCO-Stuff for ViT-S and +1.23 on VOC for ViT-L, confirming that task-specific adaptation of channel gates yields consistent improvements with minimal additional parameters. \cref{tab:upsampler_quality} reports reconstruction fidelity between upsampled and ground-truth high-resolution features using MSE ($\times 10^{3}$, $\downarrow$) and cosine similarity ($\uparrow$) across three datasets. \cref{tab:upsampler_quality} further validates the SE block: SAUCE achieves lower MSE and higher cosine similarity across all four datasets and both backbones. Reconstruction improvements scale with scene complexity; for ViT-L, MSE reduces by 3.7\% on VOC (20 classes) and 5.6\% on ADE20K (150 classes), confirming that channel calibration adapts to varying scene granularity, whereas ~\cite{jafar}, lacking this mechanism, shows degraded reconstruction quality as scene complexity increases. 
% Furthermore, these gains in reconstruction fidelity highlight the significance of improved channel calibration in the discovery of semantic masks in complex scenes, particularly in an unsupervised manner.

\subsubsection{CAFE ablation}
\begin{table}[htbp]
\centering
\caption{\textbf{CAFE component ablation on ADE20K and RSDD.} CAFE operates in three stages: (i)~Cue Discovery (CD), (ii)Attention-Guided Grouping (AGG) with optional Edge Refinement (ER), and (iii)Hierarchical Merging (HM). K-means++ replaces first 2 stages, CAFE with K-means replaces Stage-1. EASe denotes the full pipeline. Best results in \textbf{bold}, second-best baseline \underline{underlined}.}

\label{tab:ablation}
\resizebox{\linewidth}{!}{%
\setlength{\tabcolsep}{4pt}
\begin{tabular}{lccccccc}
\toprule
\rowcolor{hdrgray}
& & \multicolumn{1}{c}{\textit{Stage 1}} & \multicolumn{2}{c}{\textit{Stage 2}} & \multicolumn{1}{c}{\textit{Stage 3}} & \multicolumn{1}{c}{\textit{ADE20K}} & \multicolumn{1}{c}{\textit{RSDD}} \\
\cmidrule(lr){3-3} \cmidrule(lr){4-5} \cmidrule(lr){6-6} \cmidrule(lr){7-7} \cmidrule(lr){8-8}
\rowcolor{hdrgray}
\textit{Test configurations} & K++ & CD & AGG & ER & HM & mIoU $\uparrow$ & mIoU $\uparrow$ \\
\midrule

\rowcolor{baselineY}
(i) K-means++~\cite{10.5555/1283383.1283494} without ER          & \ding{51} & \ding{55} & \ding{55} & \ding{55} & \ding{51} & 33.5 & 38.5\\
\rowcolor{baselineY}
(ii) K-means++~\cite{10.5555/1283383.1283494}                    & \ding{51} & \ding{55} & \ding{55} & \ding{51} & \ding{51} & 40.6 & 40.5  \\
\midrule

\rowcolor{oursMint!50}
(iii) CAFE with K-means Seeding                     & \ding{51} & \ding{55} & \ding{51} & \ding{51} & \ding{51} & 45.6 & 45.1  \\
\rowcolor{oursMint!50}
(iv) CAFE without ER                   & \ding{55} & \ding{51} & \ding{51} & \ding{55} & \ding{51} & \underline{48.9} & \underline{58.1}  \\
\midrule
\rowcolor{oursMint}
EASe (Ours)                       & \ding{55} & \ding{51} & \ding{51} & \ding{51} & \ding{51} & \textbf{49.4} & \textbf{58.4}  \\
\rowcolor{oursMint}
$\Delta$                          & & & & & & \textcolor{green!60!black}{$+$15.9} & \textcolor{green!60!black}{$+$19.9} \\
\bottomrule
\end{tabular}%
}% end resizebox
\end{table}
\cref{tab:ablation} evaluates each CAFE stage by comparing it with K-means++\cite{10.5555/1283383.1283494} clustering algorithm ($K{=}200$, 20 iterations). For granularity masks we retain the Hierarchical Merging stage throughout. We evaluate five configurations: (i)~K-means++ without ER as the baseline, (ii)~K-means++ with ER, (iii)~CAFE without ER, (iv)~CAFE with K-means seeding, and (v)~the complete EASe pipeline. All variants use the same SAUCE features and are evaluated on ADE20K (150 classes) and RSDD (8 classes, fine-grained damage morphologies). The ablation confirms that each CAFE stage contributes meaningfully across both settings. Comparing configurations (i) and (iii), even without ER, CAFE surpasses refined K-means++ (ii) by an average of 17.5\% mIoU. The large ER gain on K-means++ (+7.1 on ADE20K, +2.0 on RSDD) reflects correction of noisy geometric assignments, whereas CAFE gains only +0.5 and +0.3 from the same refinement, as Stage~1's iterative prototype merging already produces semantically coherent boundaries before post-processing. Configuration (iv) improves over standalone K-means++ by +5.0 mIoU on ADE20K and +4.7 on RSDD, confirming that attention-guided grouping contributes independently of prototype initialization. The gap is most pronounced on RSDD, where EASe (v) reaches 58.4 against 38.5 for K-means++. This underscores the complementary design of EASe, SAUCE provides dense features and learned attention, CAFE exploits both for fine-grained mask discovery.

\subsubsection{Specific dataset ablation}
\begin{table}[htbp]
\centering
\caption{\textbf{Per-class IoU on Roof Subassembly Damage Detection (RSDD) dataset.} EASe vs.\ unsupervised SOTAs and the supervised ResNet34-UNet baseline~\cite{roofdamage}. $\Delta_\text{SOTA}^{U}$ is gain over best unsupervised baseline and $\Delta_\text{Sup}$ is gain over supervised. Best unsupervised in \textbf{bold}, second-best baseline \underline{underlined}.}
\label{tab:nehri_perclass}
\resizebox{\linewidth}{!}{%
\setlength{\tabcolsep}{4pt}
\begin{tabular}{l
  >{\columncolor{hdrgray}}c
  >{\columncolor{hdrgray}}c
  >{\columncolor{baselineY}}c
  >{\columncolor{tfY}}c
  >{\columncolor{tfY}}c
  >{\columncolor{tfY}}c
  >{\columncolor{oursMint}}c
  >{\columncolor{oursMint}}c
  >{\columncolor{oursMint}}c}
\toprule
\textit{Class} & \textit{Pixel \%} & \textit{Count} &
\textit{ResNet34-UNet} & \textit{DiffSeg} & \textit{DiffCut} & \textit{MaskCut} &
\textbf{\textit{EASe}} & $\Delta_\text{SOTA}^{U}$ & $\Delta_{\text{Sup}}$ \\
\midrule
Background             & 82.4 & 514 & 96.0 & \underline{82.1} & 57.8 & 81.8 & \textbf{93.6} & \textcolor{green!60!black}{+11.5} & -2.4 \\
Undamaged roof         & 11.4 & 514 & 63.0 & 18.7 & 36.3 & \underline{42.2} & \textbf{54.5} & \textcolor{green!60!black}{+12.3} & -8.5 \\
Felt underlayment      &  1.1 & 158 & 24.0 &  3.6 &  6.3 & \underline{8.1}  & \textbf{20.5} & \textcolor{green!60!black}{+12.4} & -3.5 \\
Syn.\ underlayment     &  0.9 &  85 & 53.0 & \underline{18.6} &  4.8 & 15.4 & \textbf{68.3} & \textcolor{green!60!black}{+49.7} & \textcolor{green!60!black}{+15.3} \\
Substrate              &  0.9 & 203 & 48.0 &  2.0 & 11.4 & \underline{17.3} & \textbf{50.0} & \textcolor{green!60!black}{+32.7} & \textcolor{green!60!black}{+2.0} \\
Tarp                   &  2.1 & 216 & 74.0 & \underline{18.6} &  9.3 & 15.0 & \textbf{65.5} & \textcolor{green!60!black}{+46.9} & -8.5 \\
Framing                &  1.1 & 111 & 58.0 & 23.1 & 30.0 & \underline{33.8} & \textbf{65.0} & \textcolor{green!60!black}{+31.2} & \textcolor{green!60!black}{+7.0} \\
Exposed interior       &  0.2 &  30 &  1.0 &  7.0 & \underline{13.7} & 11.2 & \textbf{50.2} & \textcolor{green!60!black}{+36.5} & \textcolor{green!60!black}{+49.2} \\
\midrule
\textbf{mIoU}          &  --  &  -- & 52.0 & 21.7 & 21.2 & \underline{28.1} & \textbf{58.4} & \textcolor{green!60!black}{+30.3} & \textcolor{green!60!black}{+6.4} \\
\bottomrule
\end{tabular}%
}% end resizebox
\end{table}
\noindent Table~\ref{tab:nehri_perclass} presents a per-class IoU breakdown comparing EASe against three unsupervised baselines (DiffSeg \cite{diffseg}, DiffCut\cite{couairon2024diffcut}, MaskCut\cite{cutler_maskcut}) and previous SOTA supervised model\cite{roofdamage} on the RSDD dataset. RSDD exhibits severe class imbalance with 82.4\% pixels covering background while the rarest class, Exposed Interior, accounts for just 0.2\%. EASe achieves 58.4 mIoU, surpassing the supervised baseline (52.0) by +6.4 points and the best prior unsupervised method, MaskCut (28.1), by +30.3 points. The supervised model nearly fails on Exposed Interior (1.0 IoU) due to extreme label scarcity, whereas EASe attains 50.2 IoU on the same class. This advantage holds consistently across low-frequency classes: Synthetic Underlayment (+15.3), Framing (+7.0), and Substrate (+2.0) over supervised, demonstrating that EASe's unsupervised mask discovery, free from label-dependent optimization, is unaffected by the class imbalance that undermines supervised training in domains where fine-grained damage annotation is scarce.

\begin{table}[htbp]
\centering
\caption{\textbf{Per-domain clIoU on OmniCrack30k dataset.} EASe vs.\ unsupervised SOTAs and the supervised nnU-Net baseline~\cite{benz2024omnicrack30k} on stricter centerline IoU metrics for suitable for evaluating crack segmentation. $\Delta_\text{SOTA}^{U}$ is gain over best unsupervised baseline. Best unsupervised in \textbf{bold}, second-best baseline \underline{underlined}.}
\label{tab:crack_domain}
\resizebox{\linewidth}{!}{%
\setlength{\tabcolsep}{4pt}
\begin{tabular}{l
  >{\columncolor{hdrgray}}c
  >{\columncolor{hdrgray}}c
  >{\columncolor{baselineY}}c
  >{\columncolor{tfY}}c
  >{\columncolor{tfY}}c
  >{\columncolor{tfY}}c
  >{\columncolor{oursMint}}c
  >{\columncolor{oursMint}}c
  >{\columncolor{oursMint}}c}
\toprule
\textit{Domains} & \textit{Pixel (in million)} & \textit{Samples} &
\textit{nnU-Net} & \textit{DiffSeg} & \textit{DiffCut} & \textit{MaskCut} &
\textbf{\textit{EASe}} & $\Delta_\text{SOTA}^{U}$ \\
\midrule
BCL             & 64.9  & 990  & 82.8 & 0.4 & \underline{1.2} & \underline{1.2} & \textbf{14.2} & \textcolor{green!60!black}{+13.0} \\
Ceramic         & 1.0   & 15   & 46.3 & 0.9 & \underline{1.5} & 0.7 & \textbf{35.9} & \textcolor{green!60!black}{+34.4} \\
CFD             & 1.5   & 10   & 85.0 & 0.7 & \underline{1.5} & 0.2  & \textbf{27.0} & \textcolor{green!60!black}{+25.5} \\
CRACK500        & 188.3 & 50   & 53.5 & 0.1 & \underline{0.2} & 0.1             & \textbf{3.0}  & \textcolor{green!60!black}{+2.8} \\
CrackTree260    & 14.0  & 26   & 92.8 & 0.6 & \underline{0.8} & 0.3             & \textbf{4.3}  & \textcolor{green!60!black}{+3.5} \\
CrsSpEE         & 68.5  & 109  & 67.0 & 0.4 & \underline{0.6} & 0.1             & \textbf{5.0}  & \textcolor{green!60!black}{+4.6} \\
CSSC            & 7.1   & 12   & 74.3 & 0.3 & \underline{0.6} & 0.4             & \textbf{6.8}  & \textcolor{green!60!black}{+6.2} \\
DeepCrack       & 6.3   & 30   & 83.4 & 0.5 & \underline{1.0} & 0.9             & \textbf{16.2} & \textcolor{green!60!black}{+15.2} \\
DIC             & 8.5   & 129  & 85.7 & 0.4 & \underline{0.7} & 0.6             & \textbf{15.7} & \textcolor{green!60!black}{+15.0} \\
GAPS384         & 8.3   & 4    & 45.5 & \underline{0.1} & \underline{0.1}  &    0.0             & \textbf{1.5}  & \textcolor{green!60!black}{+1.4} \\
Khanh11k        & 83.7  & 417  & 92.1 & 0.3 & \underline{0.6} & 0.2             & \textbf{21.0} & \textcolor{green!60!black}{+20.4} \\
LCW             & 91.1  & 87   & 10.0 & 0.2 & \underline{0.3} & 0.0             & \textbf{1.1}  & \textcolor{green!60!black}{+0.8} \\
Masonry         & 0.7   & 14   & 83.6 & 0.8 & 1.2 & \underline{2.2}             & \textbf{13.2} & \textcolor{green!60!black}{+11.0} \\
S2DS            & 91.2  & 87   & 74.3 & 0.1 & \underline{0.3} & 0.0             & \textbf{0.7}  & \textcolor{green!60!black}{+0.4} \\
TopoDS          & 84.3  & 1287 & 3.3  & 0.4 & \underline{0.9} & 0.7           & \textbf{9.7}  & \textcolor{green!60!black}{+8.8} \\
UAV75           & 2.6   & 10   & 74.6 & 0.6 & \underline{2.1} & 0.2           & \textbf{4.6}  & \textcolor{green!60!black}{+2.5} \\
\midrule
\textbf{clIoU}  &  --   &  --  & 65.9 & 0.4 & \underline{0.9} & 0.3 & \textbf{11.3} & \textcolor{green!60!black}{+10.4} \\
\bottomrule
\end{tabular}%
}% end resizebox
\end{table}
\noindent Following the official protocol in \cite{benz2024omnicrack30k}, Table~\ref{tab:crack_domain} reports per-domain centerline IoU (clIoU), a stricter skeleton-based metric than mIoU that uses dilation tolerance of 4 pixels and better suited for elongated crack structures. On OmniCrack30k, EASe consistently outperforms prior unsupervised approaches across all heterogeneous domains and achieves 11.3 clIoU (+10.4) while the best previous unsupervised baseline DiffCut (0.9), DiffSeg (0.4) and MaskCut (0.3) remain near-zero as their object-saliency focus fails to capture thin crack structures. EASe attains 35.9 (+34.4) on Ceramic, 27.0 (+25.5) on CFD, and 21.0 (+20.4) on Khanh11k, where the best unsupervised baselines remain below 2.0 IoU. Notably, EASe obtains 9.7 (+8.8) in TopoDS domain and outperfors the supervised nnU-Net (3.3) and the best unsupervised baseline, DiffCut (0.9). While certain domains such as CRACK500 and CrsSpEE exhibit single-digit performance, EASe stills leads all prior unsupervised SOTA baselines. The superior performance of EASe on both RSSD and OmniCrack30K highlights that preserving fine-grained semantic details ensures the domain-agnostic robustness necessary for segmenting the fine-grained structures.

\section{Conclusion and future works}

% For overall reference idea

We introduce Excite, Attend and Segment (EASe), a unified framework for unsupervised, domain-agnostic semantic segmentation that replaces coarse feature extraction, expensive spectral graph cuts, and image-based refinement with a single scalable architecture operating at pixel resolution. EASe significantly outperforms prior state-of-the-art while remaining efficient, and it performs strongly on both fine-grained structures and discrete objects across major standard benchmarks and domain-specific datasets with complex, non-discrete morphologies. EASe combines two components, SAUCE, a self-supervised upsampler that integrates Squeeze-and-Excitation channel calibration into cross-attention to amplify semantically useful dimensions before lifting patch tokens to pixel-level features, and CAFE, a training-free aggregator that reuses attention scores from SAUCE as a grouping signal. Our ablations show channel excitation impacts higher-dimensional backbones, where channel calibration in SAUCE shows improved advantage with scene and granularity complexity, and interative prototype merging in CAFE reduces post-hoc edge refinement while attention-guided grouping improves over conventional clustering algorithm (K-means++). Looking ahead, leveraging self-supervised learned attention as a training-free grouping signal should transfer naturally to other dense prediction tasks such as change detection and depth estimation.

\section*{Acknowledgments}
This work was performed at the University of Houston under a contract with the Commercial Smallsat Data Scientific Analysis Program of NASA (NNH22ZDA001N-CSDSA) and the NASA Decadal Survey Incubation Program: Science and Technology (NNH21ZDA001N-DSI). 

\bibliographystyle{splncs04}
\bibliography{mainv2}

\end{document}

\title{Excite, Attend and Segment (EASe): Domain-Agnostic Fine-Grained Mask Discovery with Feature Calibration and Self-Supervised Upsampling\\[6pt]
\large Supplementary Material}

\titlerunning{Excite, Attend and Segment (EASe) - Supplementary Material}

\author{}
% \authorrunning{Anonymous et al.}
\institute{}

\maketitle

% Reset section counter to start from A
\renewcommand{\thesection}{\Alph{section}}
\setcounter{section}{0}

\noindent This supplement is organised as follows. \cref{sec:supp_hm} details the hierarchical Merging (HM) algorithm. \cref{sec:supp_impl} provides implementation details including SAUCE architecture ablations (\cref{sec:supp_reduction}), per-dataset granularity with boundary-penalty parameter calibration (\cref{sec:supp_granularity}), and a channel-calibration study comparing SAUCE with JAFAR (\cref{sec:supp_upsampler}). \cref{sec:supp_qualitative} presents qualitative results across benchmarks.

\section{Hierarchical Merging Algorithm}
\label{sec:supp_hm}

This section details the Hierarchical Merging algorithm (Stage~3 of CAFE). First, we define the quantities governing boundary-gated merging, level scoring, and granularity computation, then present the complete procedure in \cref{alg:hm}.

\paragraph{Boundary-Gated Similarity.}
Given segments $i,j$ with mean prototypes $\mathbf{p}_i, \mathbf{p}_j$ from SAUCE features, the effective similarity used for merge decisions is:
\begin{equation}
  S(i,j) = \bigl[\cos(\mathbf{p}_i, \mathbf{p}_j) - \beta_{\text{bnd}}\,\hat{B}(i,j)\bigr] \cdot A(i,j),
  \label{eq:eff_sim}
\end{equation}
where $A(i,j) \in \{0,1\}$ is the 4-connected adjacency indicator, $\hat{B}(i,j)$ is the boundary strength (mean spatial gradient magnitude of SAUCE features along the shared border, normalised to $[0,1]$), and $\beta_{\text{bnd}}$ controls boundary penalty strength.

\paragraph{Level Scoring.}
After merging produces a history of $M$ snapshots, we score each level $i$ by combining the normalised next-merge cost $\hat{c}_i$ with the Calinski--Harabasz (CH) index \cite{Caliński01011974} computed in L2-normalised SAUCE feature space, measuring cluster separation:
\begin{equation}
  \mathrm{CH} = \frac{B/(K{-}1)}{W/(n{-}K)}, \quad
  B = \textstyle\sum_k n_k \|\mathbf{c}_k - \bar{\mathbf{c}}\|^2, \quad
  W = \textstyle\sum_k \sum_{\mathbf{x} \in \mathcal{C}_k} \|\mathbf{x} - \mathbf{c}_k\|^2,
  \label{eq:ch}
\end{equation}
where $K$ is the number of segments, $n$ the number of foreground pixels, $\mathbf{c}_k$ the centroid of segment $k$, $\bar{\mathbf{c}}$ the global centroid, and $\mathcal{C}_k$ the set of L2-normalised feature vectors in segment $k$. CH is defined for $2 \leq K < n$.

The combined level score is:
\begin{equation}
  s_i = (1 - \lambda)\,\hat{c}_i + \lambda\,\widehat{\mathrm{CH}}_i + \gamma_i,
  \label{eq:level_score}
\end{equation}
where $\widehat{\mathrm{CH}}_i = \mathrm{CH}_i / \max_j \mathrm{CH}_j$, $\lambda = 0.25$, and $\gamma_i$ is a persistence-gap bonus. Define the \emph{inter-level cost} as the effective similarity threshold at which level $i$ was produced, a \emph{persistence gap} occurs when this inter-level cost exceeds $3\times$ the running cumulative average of all preceding inter-level costs, indicating a large merge step. We set $\gamma_i = 0.2$ at persistence gaps and $\gamma_i = 0$ otherwise. We also define $\tau_i^{\text{floor}}$ as the minimum pairwise similarity among all merges that produced level $i$, it serves as the non-adjacent merge threshold in \cref{alg:hm}, (Line 12). The top-$N$ levels by $s_i$ are retained.

\paragraph{Granularity Scoring.}
Each retained level is annotated with scale and semantic granularity. Per segment $k$, we compute the area ratio $g_k^s = a_k / \sum_j a_j$ and the feature-range ratio $g_k^r = \psi_k / \psi_\text{fg}$, where $\psi_k = \frac{1}{C}\sum_c (f_{k,c}^{\max} - f_{k,c}^{\min})$ is the mean per-channel feature range within segment $k$ and $\psi_\text{fg}$ the same over all foreground. The level-wise scores are:
\begin{equation}
  G_\text{scale} = \frac{1}{K}\sum_k g_k^s, \qquad
  G_\text{sem} = \frac{1}{K}\sum_k \min(g_k^r,\, 1),
  \label{eq:graco}
\end{equation}

% \vfill
\begin{algorithm}[H]
\caption{Hierarchical Merging (Stage~3)}
\label{alg:hm}
\begin{algorithmic}[1]
\Require Label map $\mathbf{L}$, SAUCE features $\mathbf{F} \in \mathbb{R}^{C \times H \times W}$, threshold range $[\tau_\text{end}, \tau_\text{start}]$, step $\delta$, boundary weight $\beta_{\text{bnd}}$, min size $m$, top-$N$
\Ensure Multi-granularity hierarchy $\{L_0, \dots, L_{N-1}\}$, best level index

\State Remap $\mathbf{L}$ to compact IDs $1 \dots K$; compute spatial gradient $\nabla\mathbf{F}$
\State $\mathbf{p}_k \gets \frac{1}{|k|}\sum_{i \in k} \mathbf{F}_i$ for each segment $k$ \Comment{Segment prototypes}
\State \textbf{history} $\gets$ [$\,$]; \textbf{gaps} $\gets$ [$\,$]

\For{$\tau = \tau_\text{start}$ \textbf{down to} $\tau_\text{end}$ \textbf{by} $\delta$} \Comment{Batch sweep}
  \State Build adjacency $A$ and boundary strength $\hat{B}$ from $\mathbf{L}$
  \State $S(i,j) \gets [\cos(\mathbf{p}_i, \mathbf{p}_j) - \beta_{\text{bnd}}\,\hat{B}(i,j)] \cdot A(i,j)$
  \State Merge all pairs with $S(i,j) \geq \tau$ simultaneously via union-find; recompute prototypes as area-weighted means of merged segments
  \State Absorb segments with $< m$ pixels into most similar neighbour
  \State Record snapshot in \textbf{history}; detect persistence gaps
\EndFor

\State Score all snapshots via Eq.~\eqref{eq:level_score}; keep top-$N$ as $\{L_0, \dots, L_{N-1}\}$
\For{each retained level $L_i$} \Comment{Per-level global merge}
  \State Merge non-adjacent pairs with $\cos(\mathbf{p}_i, \mathbf{p}_j) \geq \tau_i^{\text{floor}}$
\EndFor
\State Store granularity scores (Eq.~\eqref{eq:graco}) for levels with $K > 2$
\end{algorithmic}
\end{algorithm}

% ===============================================================

\section{Implementation Details}
\label{sec:supp_impl}

\cref{tab:parameters} lists the EASe configuration parameters and their default values, which remain fixed across all datasets to produced the results shown in the main paper. SAUCE is trained self-supervised on ImageNet for 127500 steps using AdamW with a learning rate of 0.0002. Our ablation study spans backbones of varying capacity and feature dimensionality, DINOv3 ViT-S (384-d), ViT-B (768-d), and ViT-L (1024-d), as well as C-RADIOv4 (1152-d) \cite{ranzinger2026cradiov4techreport}\footnote{Code publicly released on January 27, 2026.}. The following subsections detail the SAUCE architecture ablations, the per-dataset granularity calibration, and the impact of feature channel calibration on unsupervised segmentation. 

\begin{table}[H]
\centering
\caption{\textbf{EASe configuration parameters.} SAUCE upsampler, Cue Discovery via Cross-Resolution Seeding (CRS), Attention-Guided Grouping (AGG), and Hierarchical Merging (HM).}

\label{tab:parameters}
\setlength{\tabcolsep}{5pt}
\begin{tabular}{llcl}
\toprule
\rowcolor{hdrgray}
\textit{Component} & \textit{Symbol} & \textit{Default Value} & \textit{Description} \\
\midrule
\rowcolor{baselineY}
SAUCE       & Backbone               & DINOv3-L & Default encoder \\
\rowcolor{baselineY}
            & $d_{qk}$               & 128   & Query/key projection dim \\
\rowcolor{baselineY}
            & $r$                    & 2     & SE reduction ratio \\
\midrule
\rowcolor{baselineY}
CRS         & $\tau$                 & 0.97  & Prototype merge threshold \\
\rowcolor{baselineY}
            & $T$                    & 5     & Refinement iterations \\
\midrule
\rowcolor{baselineY}
AGG         & $\alpha$               & 0.75  & Feature cost weight \\
\rowcolor{baselineY}
            & $\beta$   & 0.25  & Attention affinity weight \\
\midrule
\rowcolor{baselineY}
HM          & $N$                    & 40    & Retained hierarchy levels \\
\rowcolor{baselineY}
            & $\theta_{\text{hi}}$   & 0.99  & Sweep upper bound \\
\rowcolor{baselineY}
            & $\theta_{\text{lo}}$   & 0.30  & Sweep lower bound \\
\rowcolor{baselineY}
            & $\delta$               & 0.001 & Sweep step size \\
\rowcolor{baselineY}
            & $\beta_{\text{bnd}}$   & 0     & Boundary penalty \\
\rowcolor{baselineY}
            & $m$                    & 50    & Min segment size (pixels) \\
\bottomrule
\end{tabular}
\end{table}

\subsection{SAUCE Ablation}
\label{sec:supp_reduction}
The channel calibration in SAUCE is done by squeeze-and-excitation module, which uses a reduction factor $r$ that controls bottleneck dimensionality, the channel descriptor of dimension $C$ is compressed to $C/r$ (e.g., $1024 \to 512$ for $r{=}2$) before expansion. A smaller $r$ retains more channel information but increases parameters, a larger $r$ acts as a stronger regularizer. \cref{tab:sauce_ablation_cradio} reports the effect of varying $d_{qk}$ and $r$ on C-RADIOv4 with linear probing (SE frozen).

In SAUCE, the query encoder pools the image representation to target resolution while the key encoder downsamples to the patch grid, both project to a shared $d_{qk}$-dimensional space for cross-attention. \cref{tab:ablation_qk} varies $d_{qk}$ with fixed $r{=}2$. we observed that object-centric datasets such as ADE20K favour smaller projections ($d_{qk}{=}128$ outperforms $384$ by over 2 mIoU), while texture-heavy datasets such as RSDD favour larger $d_{qk}$, though the gain is minimal ($+0.8$). \cref{tab:ablation_se} varies $r$ with fixed $d_{qk}{=}128$, a smaller reduction ratio ($r{=}2$) consistently improves discrimination, and performance declines as $r$ increases. We default to $d_{qk}{=}128$ and $r{=}2$, which balances both regimes.

\begin{table}[H]
\centering
\caption{\textbf{SAUCE ablation study on C-RADIOv4 backbone.} (a)~Varying query-key dimension with fixed $r{=}2$. (b)~Varying SE reduction ratio with fixed $d_{qk}{=}128$.}
\label{tab:sauce_ablation_cradio}
\begin{subtable}[t]{0.48\linewidth}
\centering
\caption{Query-key dimension ($r{=}2$)}
\label{tab:ablation_qk}
\setlength{\tabcolsep}{4pt}
\begin{tabular}{c cc cc}
\toprule
\rowcolor{hdrgray}
& \multicolumn{2}{c}{\textit{ADE20K}} & \multicolumn{2}{c}{\textit{RSDD}} \\
\cmidrule(lr){2-3} \cmidrule(lr){4-5}
\rowcolor{hdrgray}
\textit{$d_{qk}$} & mIoU & Acc & mIoU & Acc \\
\midrule
\rowcolor{oursMint}
\textbf{128} & \textbf{48.60} & \textbf{78.65} & 57.92 & 95.72 \\
\rowcolor{baselineY}
256          & 46.57          & 77.93          & \underline{58.01} & \underline{95.70} \\
\rowcolor{baselineY}
384          & 46.21 & 77.76 & \textbf{58.71} & \textbf{95.79} \\
\bottomrule
\end{tabular}
\end{subtable}
\hfill
\begin{subtable}[t]{0.48\linewidth}
\centering
\caption{SE reduction ratio ($d_{qk}{=}128$)}
\label{tab:ablation_se}
\setlength{\tabcolsep}{4pt}
\begin{tabular}{c cc cc}
\toprule
\rowcolor{hdrgray}
& \multicolumn{2}{c}{\textit{ADE20K}} & \multicolumn{2}{c}{\textit{RSDD}} \\
\cmidrule(lr){2-3} \cmidrule(lr){4-5}
\rowcolor{hdrgray}
\textit{$r$} & mIoU & Acc & mIoU & Acc \\
\midrule
\rowcolor{oursMint}
\textbf{2} & \textbf{48.60} & \textbf{78.65} & \textbf{57.92} & \underline{95.72} \\
\rowcolor{baselineY}
4          & \underline{48.44} & \underline{78.58} & \underline{57.86} & \textbf{95.74} \\
\rowcolor{baselineY}
8          & 48.22          & 78.50          & 57.73          & 95.66 \\
\bottomrule
\end{tabular}
\end{subtable}
\end{table}

\subsection{Granularity Selection Across Datasets}
\label{sec:supp_granularity}

Hierarchical merging produces a fine-to-coarse sequence of label maps. The required granularity varies across target datasets depending on groundtruth annotation.To select the granularity level for evaluation, we calibrate the target granularity scores $G_{\text{scale}}$ and $G_{\text{sem}}$ on a small subset of annotated images, taking the smaller of 200 images or 10\% of the target training split. The level whose scores best match the calibrated target is selected for evaluation. This calibration step is analogous to the recursion-depth or number-of-clusters selection that prior methods (DiffCut, MaskCut) perform per dataset. EASe replaces a discrete hyperparameter search with two continuous granularity scores as a post-processing step.

\paragraph{Boundary gate ($\beta_{\text{bnd}}$).}
Hierarchical merging includes a boundary penalty $\beta_{\text{bnd}}$ in the effective similarity (\cref{eq:eff_sim}) that penalises merges across strong feature gradients. Note that $\beta_{\text{bnd}}$ is distinct from the attention affinity weight $\beta{=}0.25$ used in Stage-2 AGG (main paper, Eq.~4). For the main paper results, we set $\beta_{\text{bnd}}{=}0$ across all benchmarks. However, for domains with complex morphologies (cracks, damage boundaries, thin structures), setting $\beta_{\text{bnd}}{>}0$ preserves fragmented boundaries at coarser hierarchy levels that would otherwise be absorbed during merging.

\cref{tab:calibration} reports per-dataset granularity calibration for our default DINOv3-L backbone alongside DINOv3-B, sweeping $\beta_{\text{bnd}} \in \{0, 0.5, 1, 1.5, 2\}$ on a small calibration split to select the best configuration. \cref{tab:beta_ablation} isolates the effect of $\beta_{\text{bnd}}$ on the full validation sets with the backbone fixed to DINOv3-L, while \cref{tab:calibrated_backbone} selects both the best backbone and $\beta_{\text{bnd}}$ per dataset based on the calibration split. All main-paper results use $\beta_{\text{bnd}}{=}0$ with DINOv3-L, the calibrated results below demonstrate additional gains from per-dataset tuning.

\begin{table}[H]
\centering
\caption{\textbf{Per-dataset granularity calibration.} We sweep $\beta_{\text{bnd}} \in {0, 0.5, 1, 1.5, 2}$ across different backbones on a small calibration subset to determine the best configuration depending on granularity of the dataset. Best result per dataset in \textbf{bold}. \colorbox{gray!10}{Gray} rows denote the main-paper default (DINOv3-L, $\beta_{\text{bnd}}{=}0$). $\Delta$ mIoU compares the best $\beta_{\text{bnd}}$ with the $\beta_{\text{bnd}}{=}0$ baseline for the same backbone.}
\label{tab:calibration}
\setlength{\tabcolsep}{4pt}
\begin{tabular}{lcrcccc}
\toprule
\rowcolor{hdrgray}
\textit{Dataset} & \textit{Backbone} & $\beta_{\text{bnd}}$ & mIoU (\%) & $G_{\text{scale}}$ & $G_{\text{sem}}$ & $\Delta$ mIoU vs $\beta_{\text{bnd}}{=}0$ \\
\midrule
\rowcolor{oursMint}
ADE20K         & C-RADIOv4 & 0   & \textbf{45.76} & 0.063 & 0.430 & {-} \\
\rowcolor{baselineY}
               & DINOv3-B  & 2   & 41.74 & 0.059 & 0.435 & \textcolor{green!60!black}{+2.56} \\
\rowcolor{baselineY}
               & DINOv3-L  & 0.5 & 43.75 & 0.068 & 0.486 & \textcolor{green!60!black}{+1.05} \\
\rowcolor{gray!10}
               & DINOv3-L  & 0   & 42.70 & 0.077 & 0.511 & \\
\midrule
\rowcolor{baselineY}
Cityscapes     & C-RADIOv4 & 0   & 26.69 & 0.060 & 0.443 & {-} \\
\rowcolor{baselineY}
               & DINOv3-B  & 0.5 & 26.68 & 0.056 & 0.435 & \textcolor{green!60!black}{+1.66} \\
\rowcolor{oursMint}
               & DINOv3-L  & 1   & \textbf{29.29} & 0.104 & 0.546 & \textcolor{green!60!black}{+1.55} \\
\rowcolor{gray!10}
               & DINOv3-L  & 0   & 27.74 & 0.091 & 0.537 & \\
\midrule
\rowcolor{baselineY}
COCO-Object    & C-RADIOv4 & 2   & 47.00 & 0.093 & 0.445 & \textcolor{green!60!black}{+1.02} \\
\rowcolor{baselineY}
               & DINOv3-B  & 0.5 & 45.17 & 0.054 & 0.451 & \textcolor{green!60!black}{+1.93} \\
\rowcolor{oursMint}
               & DINOv3-L  & 1   & \textbf{47.15} & 0.168 & 0.510 & \textcolor{green!60!black}{+0.65} \\
\rowcolor{gray!10}
               & DINOv3-L  & 0   & 46.50 & 0.150 & 0.603 & \\
\midrule
\rowcolor{baselineY}
COCO-Stuff     & C-RADIOv4 & 0.5 & 50.12 & 0.178 & 0.549 & \textcolor{green!60!black}{+2.89} \\
\rowcolor{baselineY}
               & DINOv3-B  & 0.5 & 52.65 & 0.222 & 0.656 & \textcolor{green!60!black}{+4.58} \\
\rowcolor{oursMint}
               & DINOv3-L  & 1.5 & \textbf{56.07} & 0.240 & 0.676 & \textcolor{green!60!black}{+2.09} \\
\rowcolor{gray!10}
               & DINOv3-L  & 0   & 53.98 & 0.217 & 0.668 & \\
\midrule
\rowcolor{baselineY}
KITTI          & C-RADIOv4 & 1.5 & 24.21 & 0.085 & 0.531 & \textcolor{green!60!black}{+0.82} \\
\rowcolor{baselineY}
               & DINOv3-B  & 1   & 22.51 & 0.093 & 0.575 & \textcolor{green!60!black}{+0.72} \\
\rowcolor{oursMint}
               & DINOv3-L  & 1   & \textbf{25.32} & 0.115 & 0.611 & \textcolor{green!60!black}{+0.72} \\
\rowcolor{gray!10}
               & DINOv3-L  & 0   & 24.60 & 0.119 & 0.614 & \\
\midrule
\rowcolor{baselineY}
RSDD          & C-RADIOv4 & 0.5 & 56.27 & 0.013 & 0.238 & \textcolor{green!60!black}{+1.09} \\
\rowcolor{oursMint}
               & DINOv3-B  & 1   & \textbf{59.22} & 0.012 & 0.243 & \textcolor{green!60!black}{+5.85} \\
\rowcolor{baselineY}
               & DINOv3-L  & 2   & 56.96 & 0.011 & 0.239 & \textcolor{green!60!black}{+1.25} \\
\rowcolor{gray!10}
               & DINOv3-L  & 0   & 55.71 & 0.010 & 0.336 & \\
\midrule
\rowcolor{oursMint}
PartImageNet   & C-RADIOv4 & 0   & \textbf{43.65} & 0.043 & 0.399 & {-} \\
\rowcolor{baselineY}
               & DINOv3-B  & 0   & 42.32 & 0.079 & 0.527 & {-} \\
\rowcolor{gray!10}
               & DINOv3-L  & 0   & 41.32 & 0.099 & 0.541 & {-} \\
\midrule
\rowcolor{baselineY}
VOC          & C-RADIOv4 & 0.5 & 55.86 & 0.153 & 0.522 & \textcolor{green!60!black}{+2.11} \\
\rowcolor{baselineY}
               & DINOv3-B  & 0   & 58.99 & 0.134 & 0.606 & {-} \\
\rowcolor{oursMint}
               & DINOv3-L  & 0   & \textbf{63.42} & 0.206 & 0.663 & {-} \\
\bottomrule
\end{tabular}
\end{table}
\begin{table}[H]
\centering
\caption{\textbf{Per-subset granularity calibration on OmniCrack.} Best backbone and $\beta_{\text{bnd}}$ selected per subset, a dash indicates $\beta_{\text{bnd}}{=}0$ is already optimal.}
\label{tab:calibration-omnicrack}
\setlength{\tabcolsep}{4pt}
\begin{tabular}{lcrcccc}
\toprule
\rowcolor{hdrgray}
\textit{Subset} & \textit{Backbone} & $\beta_{\text{bnd}}$ & mIoU (\%) & $G_{\text{scale}}$ & $G_{\text{sem}}$ & $\Delta$ \textit{mIoU vs} $\beta_{\text{bnd}}{=}0$ \\
\midrule
\rowcolor{baselineY}
BCL              & C-RADIOv4 & 1.5 & 73.53 & 0.072 & 0.300 & \textcolor{green!60!black}{+1.29} \\
\rowcolor{baselineY}
Ceramic          & DINOv3-L  & 0.5 & 53.84 & 0.018 & 0.379 & \textcolor{green!60!black}{+0.28} \\
\rowcolor{baselineY}
CFD              & C-RADIOv4 & 0.5 & 60.87 & 0.067 & 0.362 & \textcolor{green!60!black}{+3.41} \\
\rowcolor{baselineY}
CRACK500         & C-RADIOv4 & 0.5 & 71.69 & 0.313 & 0.665 & \textcolor{green!60!black}{+6.70} \\
\rowcolor{baselineY}
CrackTree260     & C-RADIOv4 & 0.5 & 51.91 & 0.417 & 0.719 & \textcolor{green!60!black}{+0.79} \\
\rowcolor{baselineY}
CrsSpEE           & C-RADIOv4 & 0   & 73.13 & 0.018 & 0.314 & {-} \\
\rowcolor{baselineY}
CSSC             & C-RADIOv4 & 2   & 61.99 & 0.087 & 0.367 & \textcolor{green!60!black}{+4.16} \\
\rowcolor{baselineY}
DeepCrack        & C-RADIOv4 & 1.5 & 82.00 & 0.278 & 0.548 & \textcolor{green!60!black}{+0.95} \\
\rowcolor{baselineY}
DIC              & DINOv3-L  & 0.5 & 65.95 & 0.004 & 0.331 & \textcolor{green!60!black}{+7.17} \\
\rowcolor{baselineY}
GAPS384          & DINOv3-B  & 0.5 & 53.30 & 0.001 & 0.264 & \textcolor{green!60!black}{+0.86} \\
\rowcolor{baselineY}
Khanh11k         & C-RADIOv4 & 2   & 60.62 & 0.278 & 0.534 & \textcolor{green!60!black}{+8.08} \\
\rowcolor{baselineY}
LCW              & C-RADIOv4 & 0   & 52.49 & 0.000 & 0.180 & {-} \\
\rowcolor{baselineY}
Masonry          & C-RADIOv4 & 2   & 88.38 & 0.203 & 0.475 & \textcolor{green!60!black}{+7.07} \\
\rowcolor{baselineY}
S2DS             & C-RADIOv4 & 2   & 58.75 & 0.106 & 0.373 & \textcolor{green!60!black}{+4.73} \\
\rowcolor{baselineY}
TopoDS           & DINOv3-L  & 0   & 52.19 & 0.000 & 0.123 & {-} \\
\rowcolor{baselineY}
UAV75            & DINOv3-B  & 1.5 & 52.17 & 0.001 & 0.191 & \textcolor{green!60!black}{+0.17} \\
\bottomrule
\end{tabular}
\end{table}

On smaller calibration dataset, Table \ref{tab:calibration} shows the effect of increasing $\beta_{\text{bnd}}$ helps most when the backbone is weaker (DINOv3-B gains $+4.58$ on COCO-Stuff) and when datasets contain finer structures (RSDD: $+5.85$). $\beta_{\text{bnd}}{=}0$ remains optimal only on VOC and PartImageNet, where object boundaries are already well-separated. OmniCrack30K aggregates 16 crack subsets spanning diverse surface materials and imaging conditions, shown in \cref{tab:calibration-omnicrack}. Setting $\beta_{\text{bnd}} > 0$ improves 13 of 16 subsets, without boundary gating, hierarchical merging absorbs thin crack regions into the background. Among the three backbones, C-RADIOv4 achieves the best calibration-split mIoU on most subsets, with margins reaching $+8.08$ (Khanh11k) and $+7.07$ (Masonry).

\begin{table}[H]
\centering
\caption{\textbf{Effect of boundary penalty across datasets.} $\beta_{\text{bnd}}{=}0$ denotes results without boundary gating. Calibrated selects $\beta_{\text{bnd}}$ per dataset via calibration (\cref{tab:calibration}).}
\label{tab:beta_ablation}
\setlength{\tabcolsep}{4pt}
\resizebox{\linewidth}{!}{%
\begin{tabular}{llcccccc}
\toprule
\rowcolor{hdrgray}
\textit{Method} & \textit{Backbone} & $\beta_{\text{bnd}}$ & \textit{Cityscapes} & \textit{COCO-O} & \textit{COCO-S} & \textit{KITTI} & \textit{VOC} \\
\midrule
\rowcolor{baselineY}
EASe (Ours) & DINOv3-L & $0$ (fixed) & \textbf{32.8} & 43.2 & 50.9 & 36.0 & \textbf{63.6} \\
\rowcolor{oursMint}
EASe (Ours) & DINOv3-L & Calibrated & 31.9 & \textbf{44.4} & \textbf{53.3} & \textbf{36.0} & \textbf{63.6} \\
\rowcolor{oursMint}
\multicolumn{3}{l}{$\Delta$ \textit{vs} $\beta_{\text{bnd}}{=}0$} & \textcolor{black}{$-$0.9} & \textcolor{green!60!black}{$+$1.2} & \textcolor{green!60!black}{$+$2.4} & \textcolor{black}{$+$0.0} & \textcolor{black}{$+$0.0} \\
\bottomrule
\end{tabular}}%
\end{table}

\begin{table}[H]
\centering
\caption{\textbf{Effect of calibrated backbone and boundary penalty across dataset} The calibrated row uses the best backbone and $\beta_{\text{bnd}}$ selected via calibration (\cref{tab:calibration}, \cref{tab:calibration-omnicrack} ). Best result per dataset in \textbf{bold}.}
\label{tab:calibrated_backbone}
\setlength{\tabcolsep}{4pt}
\begin{tabular}{llccccc}
\toprule
\rowcolor{hdrgray}
\textit{Method} & \textit{Backbone} & $\beta_{\text{bnd}}$ & \textit{ADE20K} & \textit{PIN} & \textit{OC30K} & \textit{RSDD} \\
\midrule
\rowcolor{baselineY}
EASe (Ours) & DINOv3-L & $0$ (fixed) & 49.4 & 46.4 & 54.2 & 58.4 \\
\rowcolor{oursMint}
EASe (Ours) & Calibrated & Calibrated & \textbf{52.4} & \textbf{46.8} & \textbf{58.9} & \textbf{59.2} \\
\rowcolor{oursMint}
\multicolumn{3}{l}{$\Delta$} & \textcolor{green!60!black}{$+$3.0} & \textcolor{green!60!black}{$+$0.4} & \textcolor{green!60!black}{$+$4.7} & \textcolor{green!60!black}{$+$0.8} \\
\bottomrule
\end{tabular}
\end{table}

With a fixed backbone (\cref{tab:beta_ablation}), calibrating $\beta_{\text{bnd}}$ alone improves mIoU by up to $+2.4$ on COCO-Stuff and $+1.2$ on COCO-Objects, though Cityscapes drops slightly ($-0.9$), likely due to distributional bias in the small calibration subset.  When both backbone and $\beta_{\text{bnd}}$ are calibrated jointly (\cref{tab:calibrated_backbone}), gains are larger and consistent across all datasets: $+4.7$ on OmniCrack-30K, $+3.0$ on ADE20K, $+0.8$ on RSDD, and $+0.4$ on PartImageNet. EASe is robust to the choice of $\beta_{\text{bnd}}$ and tuning on a smaller dataset can yield additional gains, especially for domains with complex morphologies.

\subsection{Channel Calibration for Unsupervised Semantic Segmentation}
\label{sec:supp_upsampler}
\begin{table}[H]
\centering
\caption{\textbf{Effect of upsampler: SAUCE vs.\ JAFAR.} Both methods use the same DINOv3-L backbone and CAFE grouping pipeline with same configuration parameters. Best result per dataset in \textbf{bold}.}
\label{tab:upsampler_comparison}
\setlength{\tabcolsep}{4pt}
\resizebox{\linewidth}{!}{%
\begin{tabular}{lccl ccccc}
\toprule
\rowcolor{hdrgray}
\textit{Method} & \textit{SAUCE} & \textit{CAFE} & \textit{Backbone} & \textit{Cityscapes} & \textit{COCO-O} & \textit{COCO-S} & \textit{KITTI} & \textit{VOC} \\
\midrule
\rowcolor{baselineY}
JAFAR w/ CAFE & \ding{55} & \ding{51} & DINOv3-L & 31.53 & 42.40 & 52.87 & 34.79 & 62.15 \\
\rowcolor{oursMint}
EASe (Ours) & \ding{51} & \ding{51} & DINOv3-L & \textbf{31.9} & \textbf{44.4} & \textbf{53.3} & \textbf{36.0} & \textbf{63.6} \\
\rowcolor{oursMint}
\multicolumn{4}{l}{$\Delta$ \textit{}} & \textcolor{green!60!black}{$+$0.37} & \textcolor{green!60!black}{$+$2.00} & \textcolor{green!60!black}{$+$0.43} & \textcolor{green!60!black}{$+$1.21} & \textcolor{green!60!black}{$+$1.45} \\
\bottomrule
\end{tabular}}%
\end{table}
We ablate SAUCE by replacing it with the previous state-of-the-art upsampler JAFAR~\cite{jafar}, keeping all other parameters identical. \cref{tab:upsampler_comparison} reports full-validation mIoU on five benchmarks. SAUCE yields higher mIoU on all datasets with an average increase of +1.1 mIoU, confirming that SE-based channel recalibration produces more separable features for unsupervised semantic mask discovery.

% ===============================================================
\section{Qualitative Results}
\label{sec:supp_qualitative}

% We present qualitative comparisons across all evaluation benchmarks: Pascal VOC (\cref{fig:supp_qual_voc}), COCO-Stuff (\cref{fig:supp_qual_coco}), Cityscapes (\cref{fig:supp_qual_cityscapes_kitti}), ADE20K (\cref{fig:supp_qual_ade20k}), PartImageNet (\cref{fig:supp_qual_partimagenet}), OmniCrack30K (\cref{fig:supp_qual_omnicrack}), and RSDD (\cref{fig:supp_qual_nehri}). Each figure shows EASe predictions at multiple granularities alongside ground truth.
% \subsection{Domain-specialized benchmark datasets}
\subsection{Domain-specialized benchmark datasets}
\begin{figure}[!htbp]
  \centering
  \includegraphics[width=\textwidth,height=0.38\textheight]{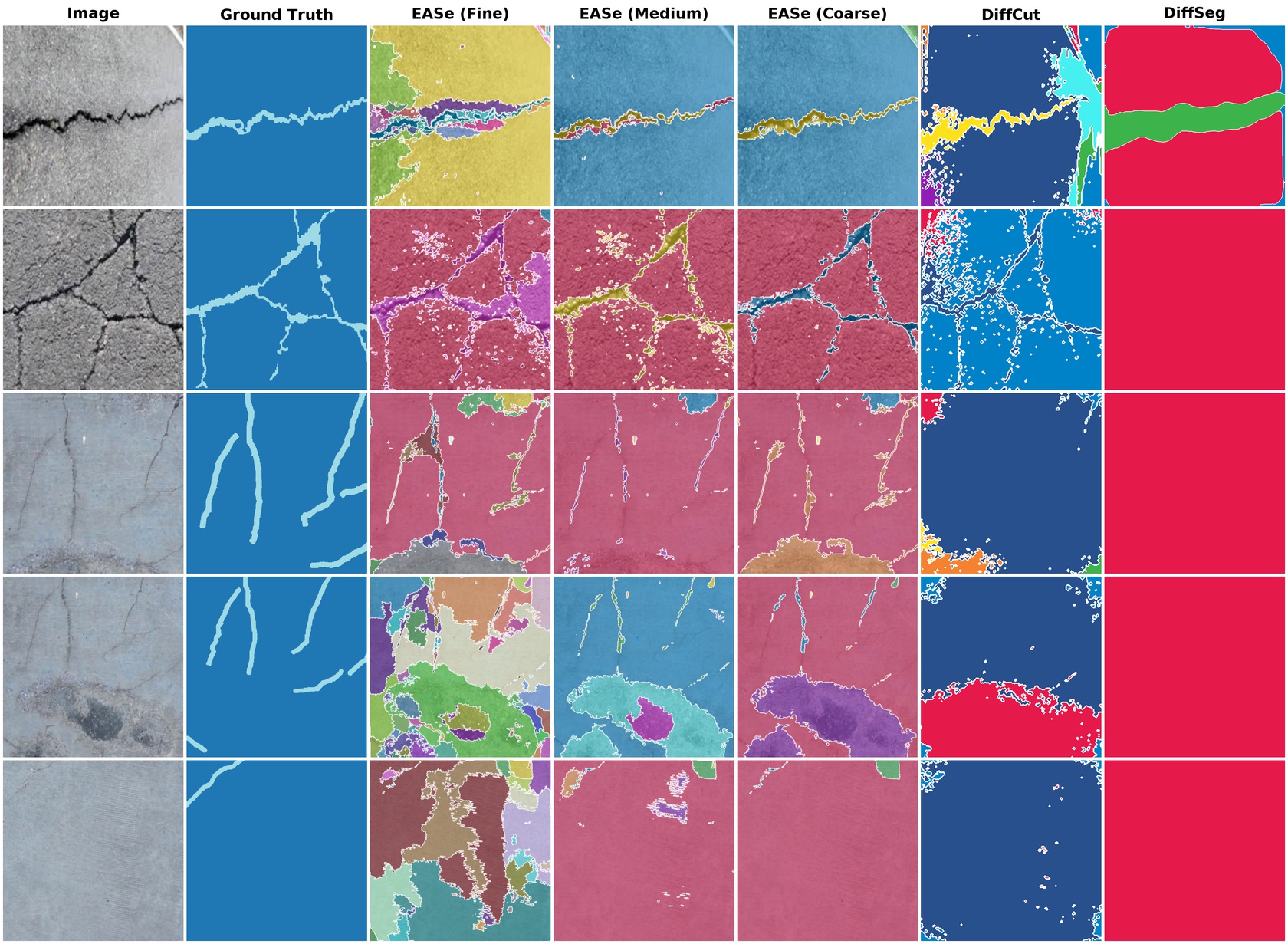}
  \caption{\textbf{OmniCrack30K.} EASe localises thin crack structures across heterogeneous surfaces and preserves connectivity for wider cracks and branching patterns. Fragmented fine hairline cracks can still be preserved.}
  \label{fig:supp_qual_omnicrack}
\end{figure}

\begin{figure}[!htbp]
  \centering
  \includegraphics[width=\textwidth,height=0.38\textheight]{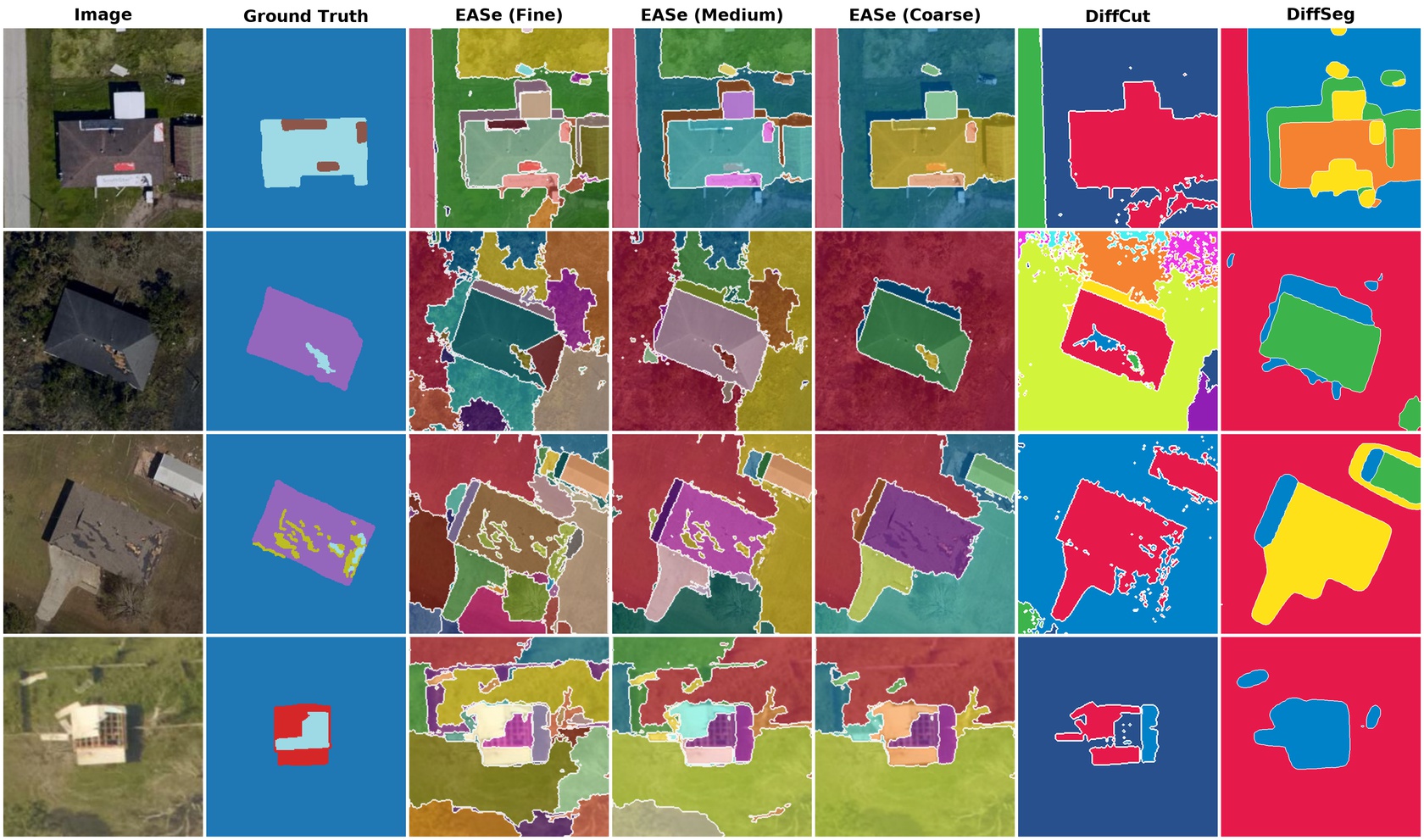}
  \caption{\textbf{RSDD.} From aerial roof imagery, EASe delineates building roof damage. Coarser levels produce regions that closely match the ground-truth annotations.}
  \label{fig:supp_qual_nehri}
\end{figure}

\noindent\cref{fig:supp_qual_omnicrack,fig:supp_qual_nehri} present qualitative results on two domain-specialized datasets. OmniCrack30K tests segmentation of thin crack structures on heterogeneous surfaces. \cref{fig:supp_qual_omnicrack} shows wider cracks and branching patterns remain well preserved, while very fine hairline cracks can still appear fragmented. Alternatively, RSDD tests segmentation of building-roof damage in aerial imagery. \cref{fig:supp_qual_nehri} shows building footprints and damaged roof structures are well-separated from surroundings, and the coarser regions align closely with the ground-truth annotations. EASe shows a consistent hierarchy across both settings. Fine-level segments capture smaller structures and local boundaries. Coarser levels merge them into larger semantic regions.

% \clearpage
% --- Page 4: RSDD ---

\FloatBarrier
\clearpage
\subsection{Standard benchmark datasets}
\begin{figure}[!htbp]
  \centering
  \includegraphics[width=\textwidth,height=0.38\textheight]{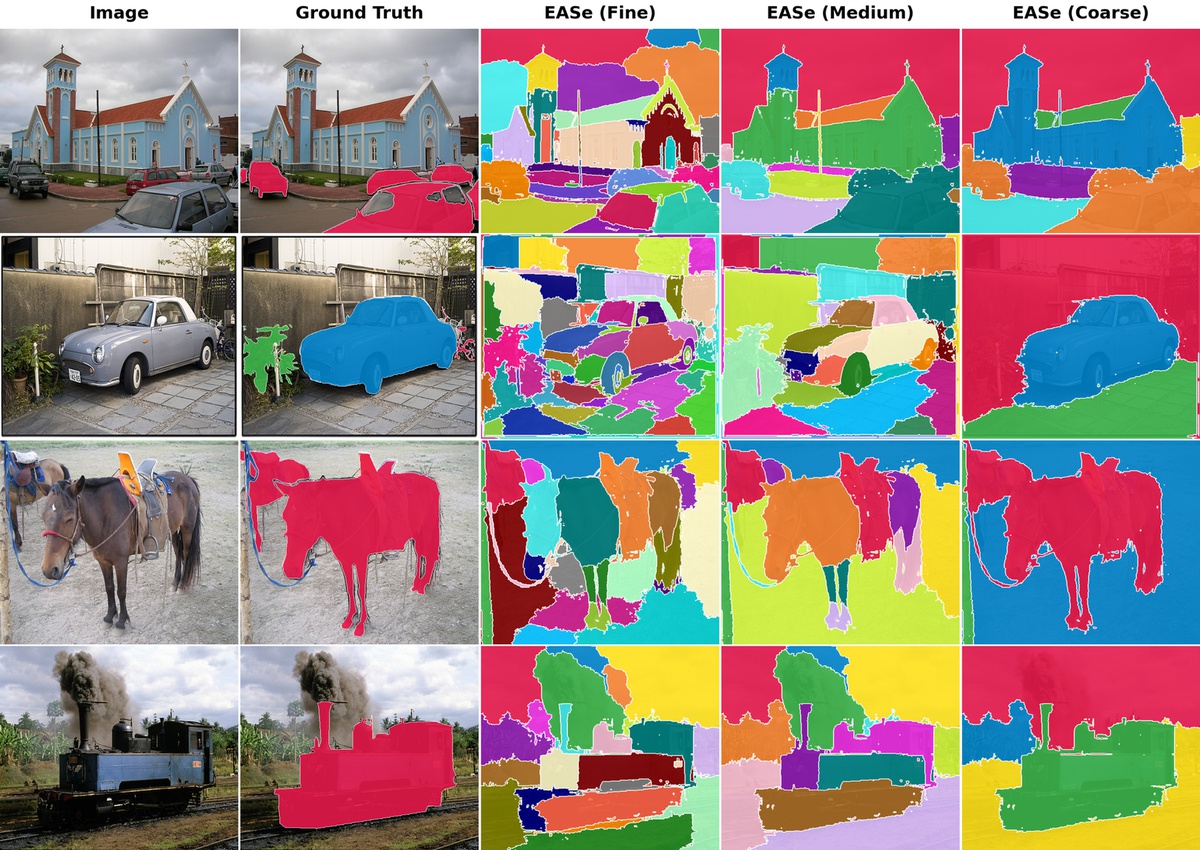}
  \caption{\textbf{Pascal VOC.} Fine-level segments trace object boundaries such as car outlines and horse legs. Coarser levels group these details into semantically coherent regions that align with the ground-truth class masks.}
  \label{fig:supp_qual_voc}
\end{figure}

\begin{figure}[!htbp]
  \centering
  \includegraphics[width=\textwidth,height=0.38\textheight]{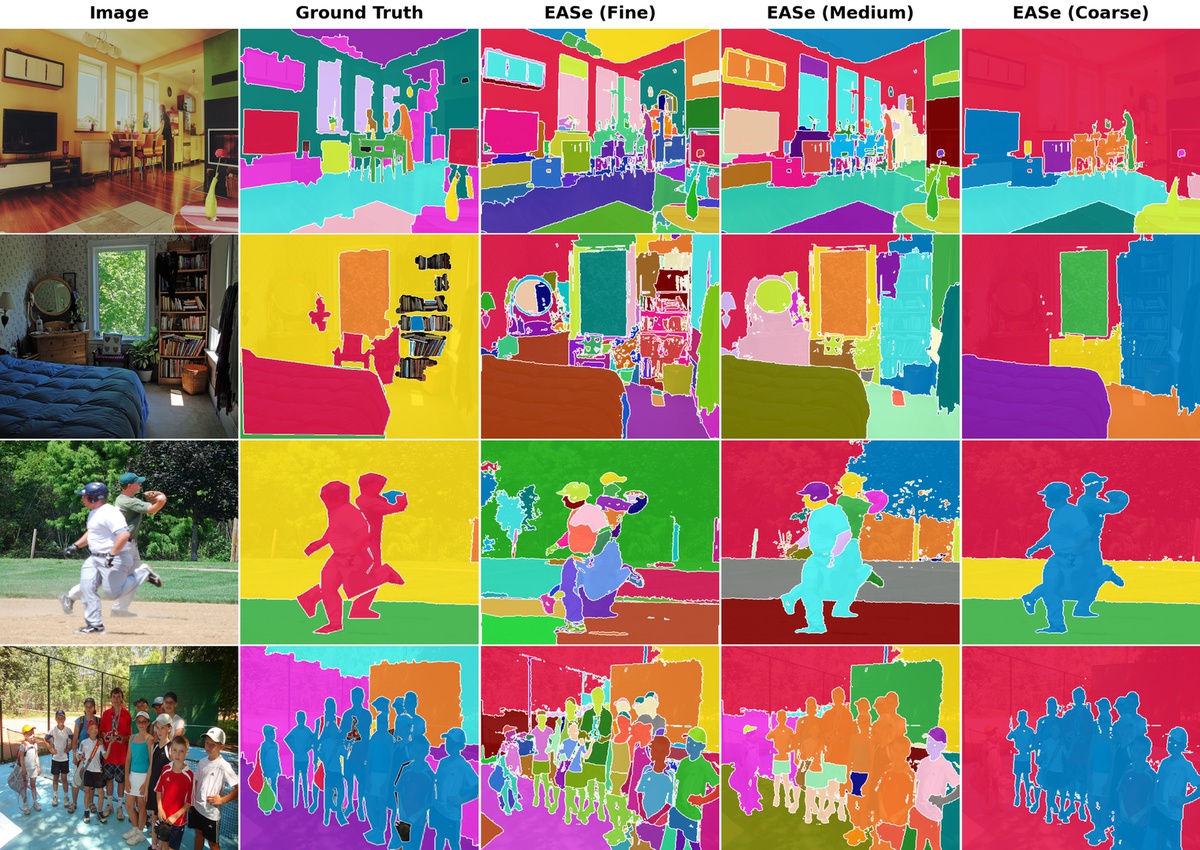}
  \caption{\textbf{COCO.} Fine-level segments separate individual objects such as furniture and people. Coarser levels group these elements into broader regions such as wall, floor, and vegetation. Occluded person instances are also clustered correctly.}
  \label{fig:supp_qual_coco}
\end{figure}

\begin{figure}[H]
  \centering
  \includegraphics[width=\textwidth,height=0.38\textheight]{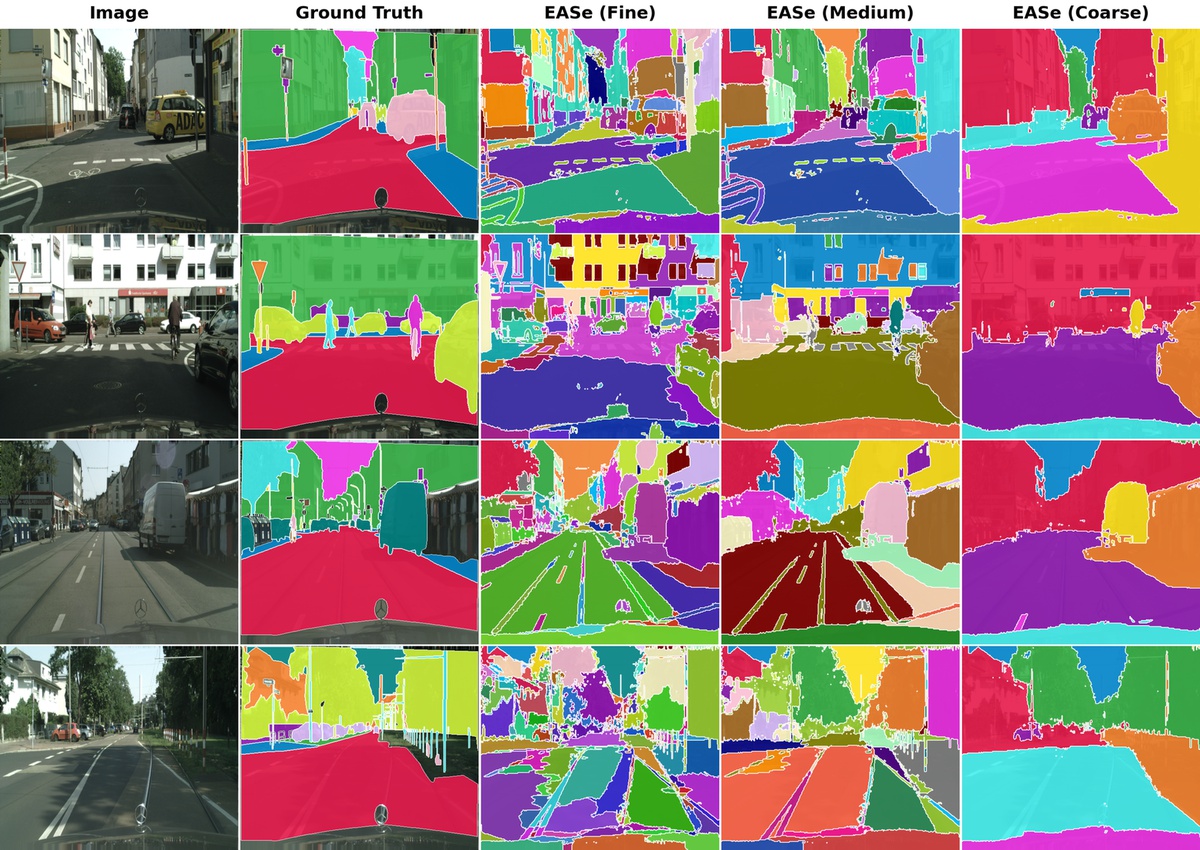}
  \caption{\textbf{Cityscapes-27.} Fine-level segments preserve sharp boundaries between road, sidewalk, and building regions. Coarser levels merge related regions while keeping sky, vegetation, and vehicles distinct.}
  \label{fig:supp_qual_cityscapes_kitti}
\end{figure}

\begin{figure}[!htbp]
  \centering
  \includegraphics[width=\textwidth,height=0.38\textheight]{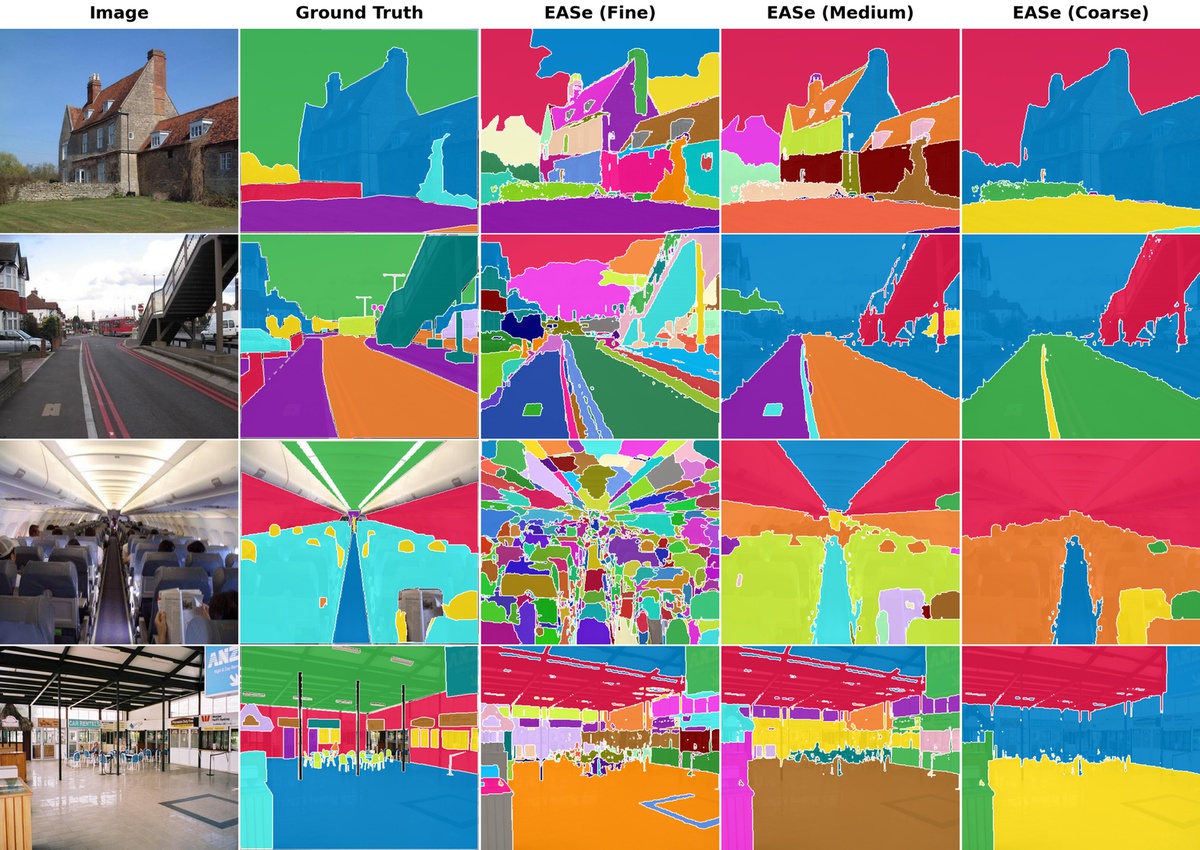}
  \caption{\textbf{ADE20K-150.} Fine-level segments resolve small structures such as windows, seat rows, and market stalls across diverse scenes. Coarser levels group larger regions such as sky and ground while preserving scene structure.}
  \label{fig:supp_qual_ade20k}
\end{figure}

\begin{figure}[!htbp]
  \centering
  \includegraphics[width=\textwidth,height=0.38\textheight]{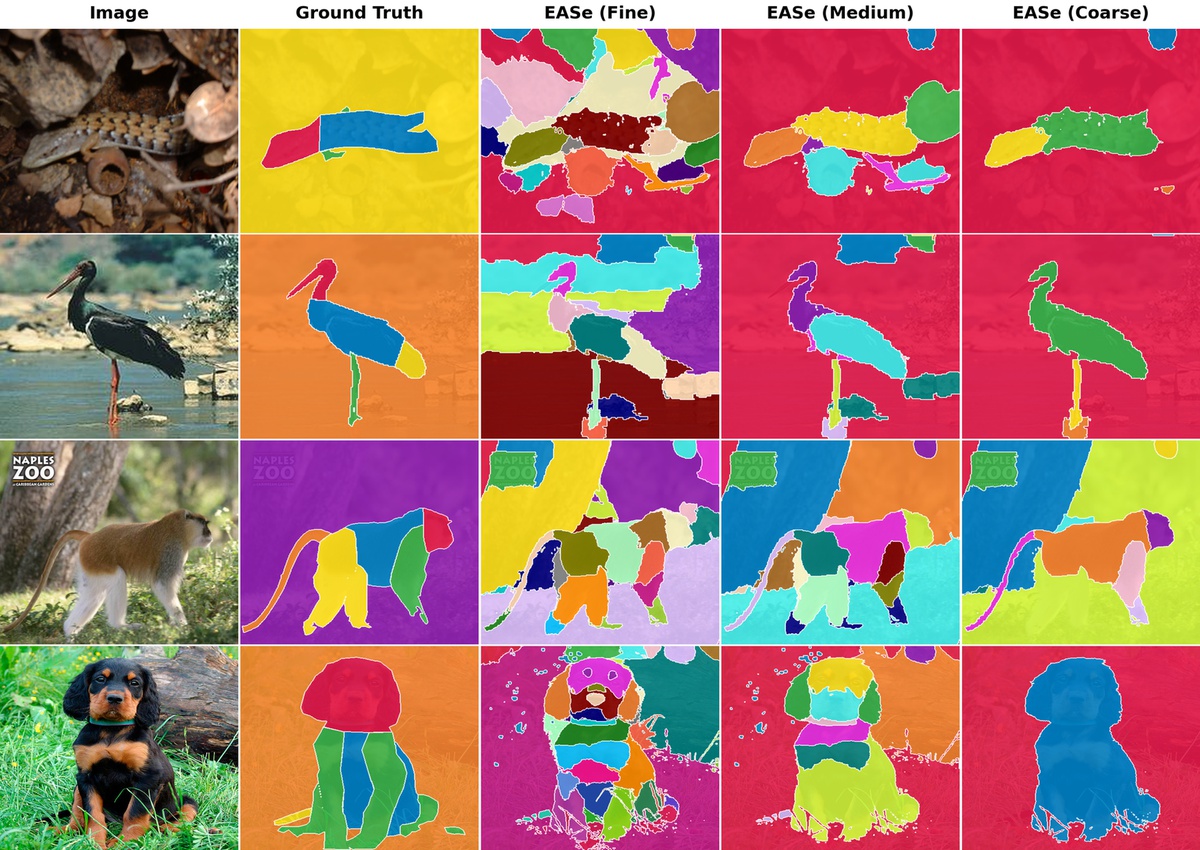}
 \caption{\textbf{PartImageNet.} Fine-level segments recover sub-object parts such as the head, legs, and tail. Coarser levels merge these parts into a single foreground object that remains clearly separated from the background.}
  \label{fig:supp_qual_partimagenet}
\end{figure}

\vspace{-2in}
\noindent\cref{fig:supp_qual_voc,fig:supp_qual_partimagenet} present qualitative results for EASe on five standard benchmark datasets. These datasets cover natural objects, scene parsing, urban imagery, part-level structure, crack detection, and aerial roof imagery. Across all examples, the hierarchy follows a consistent pattern. Fine-level segments capture smaller structures and sharper boundaries. Coarser levels group these regions into larger semantic clusters. This behavior appears across very different visual settings. On Pascal VOC (\cref{fig:supp_qual_voc}) and PartImageNet (\cref{fig:supp_qual_partimagenet}), fine segments recover object boundaries and semantic parts. On COCO (\cref{fig:supp_qual_coco}), Cityscapes-27 (\cref{fig:supp_qual_cityscapes_kitti}), and ADE20K-150 (\cref{fig:supp_qual_ade20k}), coarse levels group broad scene regions such as wall, floor, road, sky, vegetation, and buildings. These examples show a hierarchy that moves from local detail at fine granularity to broader semantic structure at coarse granularity.

% --- Page 1: VOC + COCO-Stuff ---

% \clearpage
% --- Page 2: Cityscapes + ADE20K ---

% \clearpage
% --- Page 3: PartImageNet + OmniCrack ---

\FloatBarrier
\clearpage
% ---------------------------------------------------------------
\bibliographystyle{splncs04}
\bibliography{mainv2}